\documentclass[10pt,journal,compsoc]{IEEEtran}
\usepackage{booktabs} 
\usepackage{dtklogos}
\usepackage{graphicx}
\usepackage[labelformat=simple]{subcaption}

\usepackage{algorithm}
\usepackage{algpseudocode}
\usepackage{url}
\usepackage{wrapfig}
\usepackage{diagbox}
\usepackage{mathtools}

\usepackage{array}
\newcolumntype{C}[1]{>{\centering\let\newline\\\arraybackslash\hspace{0pt}}m{#1}}

\newtheorem{definition}{Definition}
\newtheorem{lemma}{Lemma}

\DeclareMathOperator*{\argmin}{argmin}
\DeclareMathOperator*{\argmax}{argmax}
\newenvironment*{proof}{\paragraph*{\textbf{Proof}}}{\hfill$\square$}

\ifCLASSOPTIONcompsoc
  \usepackage[nocompress]{cite}
\else
  \usepackage{cite}
\fi

\ifCLASSINFOpdf
\else
\fi

\usepackage{amsmath}
\usepackage{amsfonts}
\usepackage{amssymb}

\begin{document}
\onecolumn
\textbf{NOTE: }This paper is a preprint to submission accepted in IEEE Transactions on Knowledge and Data Engineering ($DOI:10.1109/TKDE.2019.2911681$). © 2019 IEEE.  Personal use of this material is permitted.  Permission from IEEE must be obtained for all other uses, in any current or future media, including reprinting/republishing this material for advertising or promotional purposes, creating new collective works, for resale or redistribution to servers or lists, or reuse of any copyrighted component of this work in other works. 
\twocolumn
\pagebreak

%
\title{Mining Novel Multivariate Relationships in Time Series Data Using Correlation Networks}

\author{Saurabh~Agrawal,
        Michael~Steinbach,
        Daniel~Boley,
        Snigdhansu~Chatterjee,
        Gowtham~Atluri,
        Anh~The~Dang,        
        Stefan~Liess,
        and Vipin~Kumar 
\IEEEcompsocitemizethanks{\IEEEcompsocthanksitem Saurabh Agrawal, Michael Steinbach, Daniel Boley, Snigdhanshu Chatterjee, Stefan Liess, and Vipin Kumar, are with University of Minnesota, Minneapolis, MN.\protect\\
E-mail: [agraw066,stei0062,boley,chatt019,liess,kumar001]@umn.edu
\IEEEcompsocthanksitem{Gowtham Atluri and  Anh The Dang are with University of Cincinnati,Cincinnati, OH. \protect\\ E-mail: atlurigm@ucmail.uc.edu}

}
} 

\markboth{IEEE Transcations on Knowledge and Data Engineering}%
{Shell \MakeLowercase{\textit{et al.}}: Bare Demo of IEEEtran.cls for Computer Society Journals}

\IEEEtitleabstractindextext{%
\begin{abstract} \small\baselineskip=9pt
In many domains, there is significant interest in capturing novel relationships between time series that represent activities recorded at different nodes of a highly complex system. In this paper, we introduce multipoles, a novel class of linear relationships between more than two time series. A multipole is a set of time series that have strong linear dependence among themselves, with the requirement that each time series makes a significant contribution to the linear dependence. We demonstrate that most interesting multipoles can be identified as cliques of negative correlations in a correlation network. Such cliques are typically rare in a real-world correlation network, which allows us to find almost all multipoles efficiently using a clique-enumeration approach. Using our proposed framework, we demonstrate the utility of multipoles in discovering new physical phenomena in two scientific domains: climate science and neuroscience. In particular, we discovered several multipole relationships that are reproducible in multiple other independent datasets and lead to novel domain insights.
\end{abstract}

\begin{IEEEkeywords}
multivariate linear patterns; correlation mining; spatio-temporal; climate teleconnections; fMRI
\end{IEEEkeywords}}

\maketitle
\IEEEpubidadjcol

\IEEEdisplaynontitleabstractindextext
\IEEEpeerreviewmaketitle

\IEEEraisesectionheading{\section{Introduction}\label{sec:Intro}}
\IEEEPARstart{I}{n} many domains, understanding the relationships between time series is essential for obtaining actionable insights. For instance, in climate science, pressure dipoles, which are pairs of locations with strong negative correlations in their Sea Level Pressure time series, have been extensively studied, and have been linked with anomalous weather events all over the globe such as forest fires, hurricanes etc.\cite{kawale2013graph,taschetto2014cold,wallace1981teleconnections}. Similarly, in neuroscience, researchers have discovered pairs of brain regions that exhibit positive correlations between their activity time series. These correlated time series represent brain regions exhibiting synergistic activity \cite{atluri2016brain}.


In this work, we define a novel class of linear relationships, called \textbf{multipoles}, that involve more than two time series, also referred to as variables.  We say that a set of variables is a multipole if i) the variables show strong linear dependence, and ii) each variable makes a significant contribution to the linear dependence, i.e., excluding any of the variables from the set significantly weakens the strength of the linear dependence among the remaining variables. We define linear dependence in terms of the variance of a linear combination of the standardized (zero mean, unit variance) vectors. We measure the strength of the linear dependence of a set of time series by the variance of their \textit{least variant} linear combination, i.e., a (unit-length) linear combination that has the minimum variance. The smaller the variance of this linear combination, the more constant (less variable) the linear combination, and thus higher the linear dependence, and vice-versa. We define the contribution of each variable to the linear dependence as the reduction in the strength of the linear dependence of the smaller set when this variable is removed. We define linear gain as the minimum contribution of included variables to the linear dependence.   



\begin{figure}[t]
\centering
\includegraphics[width = 5cm,height=3cm]{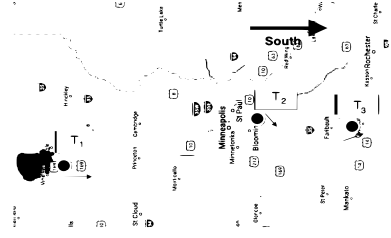} 
\caption{A multipole among daily traffic time series $T_1$,$T_2$, and $T_3$ observed for three roads in Minneapolis}
\label{fig:good_map}
\end{figure}

\begin{figure}[tb]
\centering
\begin{subfigure}[t]{0.24\textwidth}
\includegraphics[width=\textwidth,height = 7.5cm]{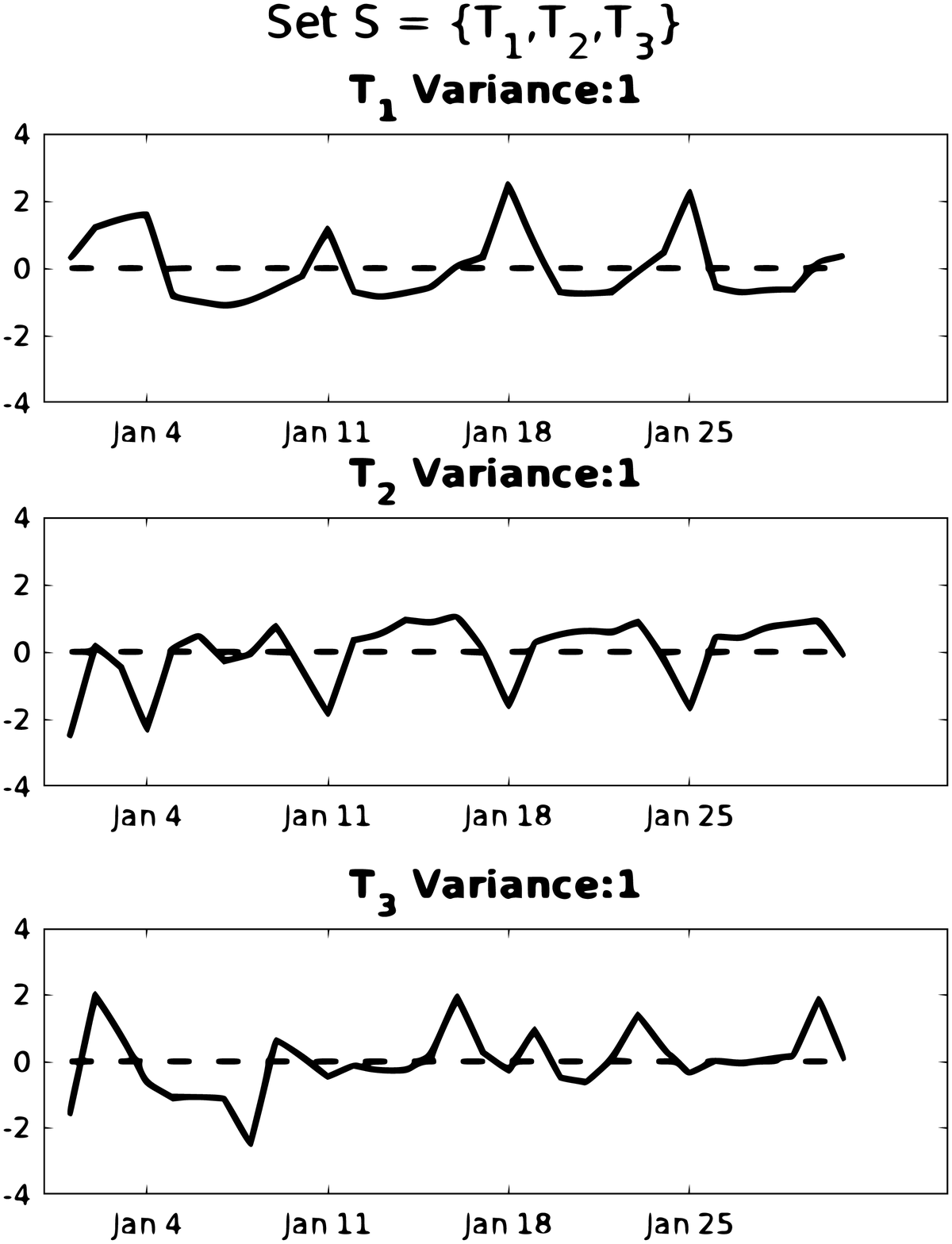}
\centering
\caption{}
\label{fig:rawts}
\end{subfigure}
\centering
\begin{subfigure}[t]{0.24\textwidth}
\includegraphics[width=\textwidth,height = 7.5cm]{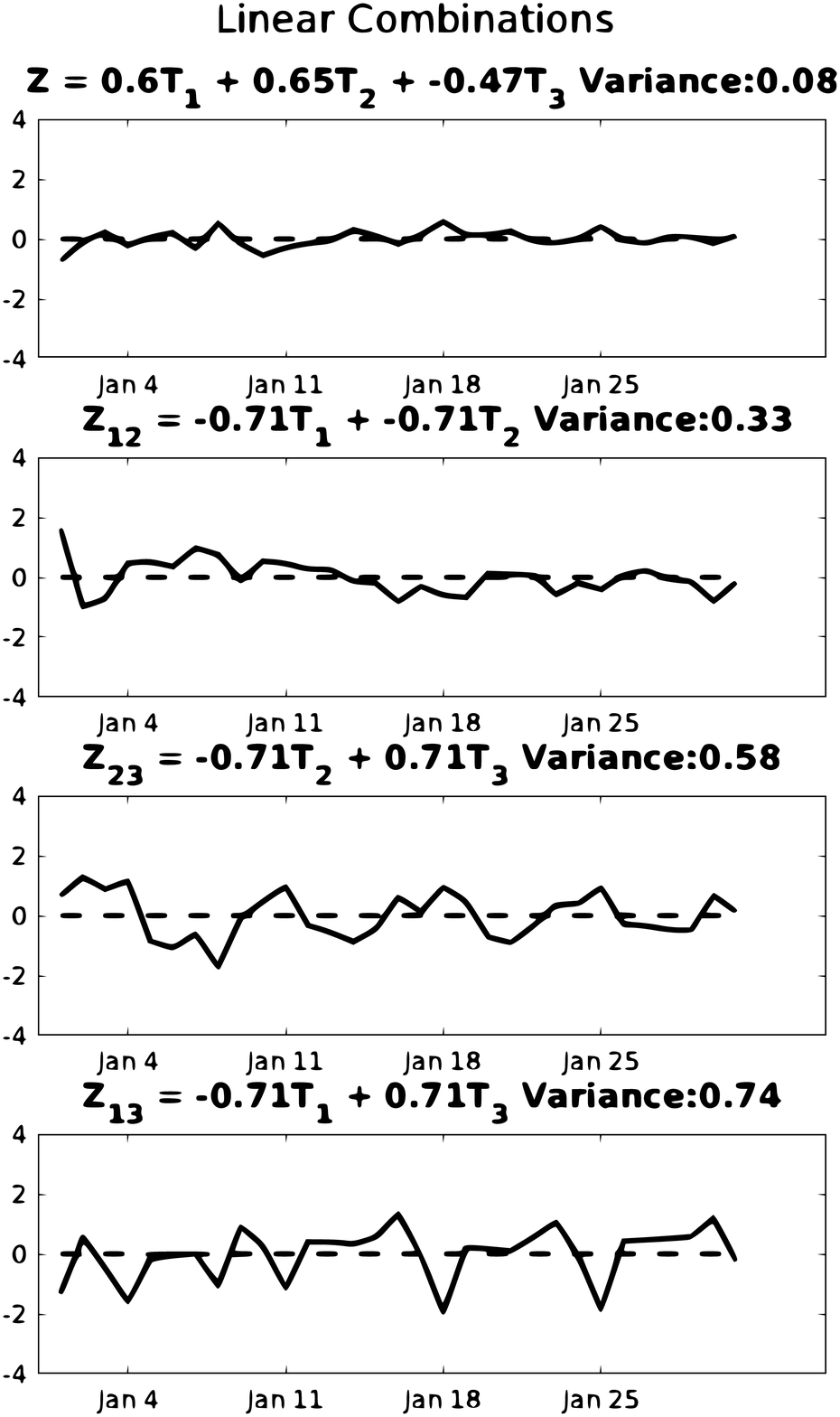}
\centering
\caption{}
\label{fig:lc}
\end{subfigure}
\caption{(a) Standardized (zero mean and unit variance) daily traffic volume for the three roads shown in Figure~\ref{fig:good_map}. (b) Time series of their linear combinations}
\label{fig:good_tripole}
\end{figure}

We next illustrate multipoles with a real-world example. Consider a set $S$ of three time series $T_1$, $T_2$, and $T_3$, shown in Figure~\ref{fig:rawts}, which capture the traffic volume on three different roads in Minnesota as shown in Figure~\ref{fig:good_map}. The three bottommost plots of Figure~\ref{fig:lc} show the least variant linear combination for each possible pair, while the top plot of Figure~\ref{fig:lc} shows the least variant linear combination for all three. Note that the linear combination of all three time series has a variance of only $0.08$.Thus, we say that there exists a strong, although not perfect, linear dependence among the above three time series.

Further, note that each of the three time series forms a crucial component of the relationship, as excluding any one of them will significantly weaken the strength of the linear dependence among the remaining two time series. For example, if $T_3$ is excluded from the set, the variance of the least variant linear combination $Z_{12}$ of $T_1$ and $T_2$ turns out to be $0.33$, which is much higher than the variance $0.08$ of $Z$, thus indicating a significant contribution from $T_3$. Similarly, if we exclude $T_1$ or $T_2$ instead of $T_3$, the variance of least variant combinations become $0.58$ and $0.74$. (See the three bottommost plots of Figure~\ref{fig:lc}.) The linear gain for $\{T_1, T_2, T_3\}$ is $\min\{0.33, 0.58, 0.74\} - 0.08 = 0.25$.

An explanation for the relationships illustrated above can be provided by the notion of conservation of flow. Two of the time series, $T_1$ and $T_2$, were observed on the roads that act as major tributaries to the highway where $T_3$ is being observed. All the southbound traffic coming from the tributaries is likely to merge at the highway, thus leading to a strong linear dependence among the three time series. Omitting any single time series leads to a weaker linear dependence because, unlike the highway, traffic flow in the tributaries significantly changes during weekends (see peaks and drops every Sunday), which weakens the strength of their pairwise linear relationships with highway. However, the simultaneous rise and fall in traffic on both tributaries complements each other, resulting in a stronger linear dependence among the three time series.

Multipoles can also be seen in other domains. For instance, we used the techniques presented in this paper to find several previously unknown multipole relationships between climate variables observed at more than two distant locations. Similarly, in neuroscience, we found novel multipole relationships between different brain regions that are triggered by specific visual and auditory stimuli. Some of these relationships were found to have insightful domain interpretations. (See  Section~\ref{Sec:PhysInt}.)

The relationships and conditional independence of two or more variables have been widely studied in different contexts using a variety of techniques including regression models \cite{tibshirani1996regression,lozano2009spatial,chatterjee2012sparse,carroll2006measurement}, PCA-based approaches\cite{barnston1987classification,ding2005circumglobal}, structure learning methods \cite{friedman2008sparse,meinshausen2006high}, correlation network analysis \cite{tsonis2011community,donges2009backbone,kawale2011discovering,agrawal2017tripoles}, etc. However, as discussed in Section \ref{sec:RelWork}, multipole relationships as defined here can be viewed as novel in the sense that none of these previous approaches can be used to find the type of relationship represented by multipoles in the data. 


A na\"{\i}ve approach to find all multipoles in a time series dataset would be to enumerate all possible combinations of time series and measure the strength of their linear dependence and linear gain. However, this is computationally infeasible due to the combinatorial nature of the search space. In this paper, we present an efficient approach to find multipoles.


This approach formulates the multipole-search problem as a clique-enumeration problem in a correlation network, where each node represents a time series, and the weight of an edge between two nodes represents the strength of the linear correlation between the corresponding time series. Our proposed problem formulation is motivated by the following two key empirical observations: the upper limit on the linear gain of a multipole is dependent on  i) the size of multipole, and ii) the maximum correlation strength among two variables in a multipole. 

Leveraging these observations, we propose a novel \textbf{C}lique Based \textbf{M}ultipole S\textbf{e}arch (CoMEt) approach to find most interesting multipoles in a time series dataset. The central idea of the approach is to identify and restrict the search for multipoles to family of subsets, which we refer to as `promising candidates` that are more likely to exhibit multipole relationships with stronger linear gain between their members. Using the above empirical observations,  we show that most promising candidates of multipoles with high linear gain appear in a correlation network either as \emph{negative cliques}, i.e. sets with all pairwise correlations being negative, or \emph{negative-equivalent cliques}, that can be transformed into negative cliques by flipping the signs of one or more of its member variables (see Definitions~\ref{def:negcl} and \ref{def:pseudo} for further details). The number of promising candidates typically turns out to be much smaller in scenarios where multipoles with high linear gain are desired, thereby contributing to the remarkably high computational efficiency of our approach, although with some loss of completeness in the final output. 

Furthermore, we propose CoMEtExtended, a more generalized version of CoMEt approach, where we redefine what constitutes a promising candidate. In particular, CoMEtExtended involves an additional parameter that can be tuned to expand or prune the scope of promising candidates beyond negative cliques and negative-equivalent cliques, which allows one to achieve a better trade-off between computational efficiency and completeness at different thresholds of linear gain (see Section~\ref{Sec:COMETEXT} for further details).

Our paper makes several key contributions: 1) We formally define multipole, a novel relationship in time series data and devise measures to quantify its interestingness. 2) We formulate a novel and computationally efficient pattern-mining approach to find most interesting multipoles in a time series dataset. 3) Further, we propose an empirical framework to evaluate discovered multipoles that includes an empirical procedure to assess the statistical significance of multipoles. 4) Using our proposed framework, we demonstrate the relevance of multipoles to two scientific domains: climate science and neuroscience.

\section{Definitions and Notations}\label{Sec:Def}
Let $S = \{X_1,X_2,...,X_k\}$ denote a set of $k$ standardized (zero mean, unit variance) time series observed over $T$ consecutive timestamps. (We will use the terms `time series` and `variables` interchangeably in what follows.)  Also, let $\textbf{X}$ be the corresponding $T \times k$ data matrix and let $\Sigma = \textbf{X}^T \textbf{X}$ be the $k \times k$ covariance matrix. Since all the variables (time series) are standardized, the correlation matrix and covariance matrices will be exactly the same for $X$.  We next define a few measures on a set of variables which will be eventually used to formally define a multipole.



\begin{definition}\label{Def:NLC}
A \textit{Normalized Linear Combination }(\textbf{NLC}) refers to a linear combination with normalized weights. Specifically, for a given vector $l \in \mathbb{R}^k$ with $||l||_2 = 1$, a normalized linear combination of variables in the set, $S$, is given as $Z_S = \mathbf{X}l$. 
\end{definition}


\begin{definition}\label{Def:CLC}
Given a set of variables $S = \{X_1,X_2,...,X_k\}$, let $Z^*_{S}$ be the \textbf{Least Variant Normalized Linear Combination} (LVNLC) of variables, i.e., the NLC that has the least variance across the $T$ observations. Formally, \[Z^*_{S} = \mathbf{X} l^*~~~\mathrm{where}~~ l^* = \argmin\limits_{l \in \mathbb{R}^k,||l||_2 = 1} var(\mathbf{X} l). \]
\end{definition}



In linear algebra, a set $S =$ $\{$ $X_1$, $X_2$, ..., $X_k\}$ of variables is said to be linearly dependent if there exists a linear combination of the variables that equals the 0 vector. Since we have subtracted the mean from our vectors (time series), any linear combination will also have a mean of zero, and thus the variance of a linear combination of $S$ is equivalent to finding the L2 norm of the linear combination. Hence, the LVNLC of $S$ finds the linear combination of $S$ that is the closest approximation to the 0 vector in terms of the L2 norm. If the linear combination of our time series (vectors) is perfectly constant, i.e., the time series vectors are linearly dependent, the variance of LVNLC will be exactly zero. In the other extreme case, when all the variables are mutually orthogonal to each other, the variance of LVNLC will be equal to 1. (This is because we have standardized the times series in $S$ to have unit variance.) Thus, variance of LVNLC can be used as an inverse indicator of the strength of linear dependence.  Based on this observation, we define the linear dependence of a set as follows:


\begin{definition}[\textbf{Linear Dependence:}] The linear dependence, $\sigma_S$, of a set of vectors $S$ is given by $1-var(Z^*_{S})$. 
\end{definition}
\textbf{Relation between $\sigma_S$ and least eigenvalue of correlation matrix:} Notice that on performing the eigenvalue decomposition of correlation matrix $\Sigma$ of the variables in the set, the eigenvalues so obtained are equal to the variances of the projections of the data along their corresponding eigenvectors. Since $Z^*_{S}$ corresponds to the direction of least variance, $Z^*_{S}$ is nothing but the eigenvector corresponding to the least eigenvalue $\lambda_{min}$ of $\Sigma$. Thus, the variance of $Z^*_{S}$ is exactly equal to the least eigenvalue of $\Sigma$ and therefore, 
\begin{align}\label{Eqeig}
\sigma_S = 1 - \lambda_{min}
\end{align}

Before proceeding further, the following two properties of linear dependence are noteworthy:
\begin{lemma}\label{lem:ld_range}
For any set $S$ of standardized variables, $\sigma_S \in [0,1]$.
\end{lemma}
\begin{proof}
See the Supplemental material.
\end{proof}





\begin{lemma}\label{lem:ld}
The linear dependence of a set $S$ is always less than or equal to that of its supersets.
\end{lemma}
\begin{proof}
See the Supplemental material.
\end{proof}

Although linear dependence indicates a strong relationship among the variables, it does not exclude the presence of irrelevant variables in the set. For instance, let $S = \{X_1,X_2,X_3,X_4\}$ be a set of linearly dependent variables with the linear relation being $X_1 + X_2 + X_3 +0X_4= 0$. Although the four variables are linearly dependent, $X_4$ is an irrelevant variable and can be pruned from $S$ without weakening the linear dependence among remaining variables. Hence, to avoid irrelevant variables in the pattern, we next propose a measure called \emph{linear gain} that checks the minimum contribution from all member variables to the linear dependence of the set.

\begin{definition}[\textbf{Linear Gain:}]\label{Def:DelG}
The linear gain of a set $S$ with $|S|>2$ is measured as the gain in the linear dependence of $S$ with respect to one of its proper subsets $S'$ that has strongest linear dependence. Mathematically, we can write linear gain of $S$ as 
\begin{align}\label{Eq:gain1}
\Delta\sigma_S =  \sigma_S- \max\limits_{S'\subset S}\sigma_{S'}
\end{align}
\end{definition}

From Lemma~\ref{lem:ld}, we get that the linear dependence of a set is always greater than that of its subsets, which implies that the linear gain of a set will always be positive. Furthermore, we can say that the subset with strongest linear dependence will be of size $|S|-1$. Thus, the linear gain can be more precisely written as

\begin{align}\label{Eq:gain2}
\Delta\sigma_S =  \sigma_S- \max\limits_{X_i \in S}\sigma_{S-X_{i}}
\end{align}

Higher values of linear gain imply that a more significant drop in linear dependence would be observed if any one of the variables are excluded from the set, thereby ensuring that no irrelevant variables are included. Furthermore, a high threshold on linear gain will avoid redundancies in the set. For instance, in the  example from traffic data described in previous section, suppose we insert $T_4$ into the set $\{T_1,T_2,T_3\}$, where $T_4$ comes from a sensor close to the sensor for $T_3$, on the the same road. Since $T_4$ is almost a duplicate of $T_3$, then the linear dependence of the resultant set  $S' = \{T_1,T_2,T_3,T_4\}$ would be almost $1$, since there would exist a linear combination $T_4 - T_3 \approx 0$ with near perfect linear dependence. However, that would also imply that many of its subsets, e.g. $\{T_1,T_3,T_4\}$, would have near perfect linear dependence. Hence by definition, the linear gain of $S'$ will be very close to zero. More generally, a high threshold on linear gain will also avoid multicollinearity in the set. For instance, consider a set $S = S_1 \cup S_2 $ that consists of two independent subsets $S_1$ and $S_2$ of perfectly linearly dependent variables. By definition, the linear gain of such a set $S$ will be 0, and hence will be discarded.

Using the above definitions, we next present the formal definition of a multipole. 
\begin{definition}\label{Def:Mul}
A \textbf{multipole} refers to the set $S$ of variables with $|S|\geq 2$ such that $\sigma_S \geq \sigma $ and $\Delta\sigma_S \geq \delta$, where $\sigma$ and $\delta$ are user-specified thresholds. 
\end{definition}


We next define the notion of maximality in a multipole. 
\begin{definition}[\textbf{Maximal Multipole}]\label{Def:MultMaximal}
In a set $Q$ of multipoles, a multipole $S$ is considered to be \textbf{maximal} if none of its supersets are in $Q$. 
\end{definition}

A maximal multipole is likely to capture the underlying signal more comprehensively compared to its subsets. Hence, all non-maximal multipoles could potentially be pruned in the final output of the search. 

Using the above definitions, we formulate the multipole-discovery problem as the following:

\begin{definition}[\textbf{Problem Formulation:}]\label{Def:Prob}
Given $\delta$ and $\sigma$, find  the set $P$ of all maximal multipoles in a given time series dataset. 
\end{definition}

\section{Related Work}\label{sec:RelWork}
In this section, we present an overview of different techniques that have been applied to study relationships between two or more variables and discuss their similarities and differences with multipoles. 

\textbf{Eigenanalysis-based Approaches:} Eigenanalysis based-approaches such as Principal Component Analysis (PCA) \cite{barnston1987classification,ding2005circumglobal} and Independent Component Analysis \cite{van2004functional} commonly focus on finding linear combinations of variables that capture the dominant global signals of variability in the data. Thus, they are interested in the largest eigenvalues and often treat the smaller eigenvalues as noise and discard them. If we consider a dataset of the traffic time series at all roads in the city and apply PCA on it, it is expected to capture dominant global patterns of variability such as work-home traffic patterns, patterns influenced by social events, etc. In contrast, multipoles capture the least-variant local linear dependencies among small subsets of vectors. (Roads in this example.) This corresponds to subsets of vectors whose least eigenvalue (of the correlation matrix) is small, i.e., close to 0. 

\textbf{Regression Models:} Regression models, such as ordinary least squares (OLS) and its regularized variants , e.g. LASSO \cite{tibshirani1996regression}, are used to find a linear combination of independent variables that predicts the given dependent variable with high accuracy \cite{lozano2009spatial,chatterjee2012sparse}. The independent variables can therefore be considered as showing a strong linear dependence with the dependent variable. However, such techniques do not have a notion of gain and thus are not designed to find multipole relationships. To illustrate, consider once again the highway time series example. If we use LASSO to find a set of predictors for the highway time series $T_1$, LASSO would always include the time series that is most strongly correlated with $T_1$ as the predictor, say $T_1'$, collected at another sensor on the same highway. Consequently, it will miss all multipole patterns that include $T_1$ but not $T_1'$, many of which otherwise could be capturing interesting and non-trivial relationships of $T_1$ with distant road stations.

\textbf{Error-In-Variables Models:} Error-in-Variables (EIV) models are a special class of regression models that account for uncertainties in the measurements of both dependent and independent variables (unlike standard regression models, which assume that independent variables are measured accurately). Like EIV models, the definition of multipoles does not create any distinctions among the participating variables. One of the multivariable linear EIV models, named Total Least Squares (TLS) has striking similarities with the proposed definition of multipoles \cite{carroll2006measurement}. In particular, TLS focuses on learning a linear combination of a given set of dependent and independent variables to minimize the joint residual error in all the variables, which is exactly same as finding a linear combination of variables with highest linear dependence. Like multipoles, the solution to TLS  is obtained by computing the eigenvector corresponding to the least eigenvalue of the covariance matrix of the given set of variables. TLS does not have a notion of linear gain, although it can be shown that a given set of variables obtains a high linear gain only if 1) TLS obtains a unique solution, and 2) all the regression coefficients obtained in the solution of TLS are significantly higher than zero in magnitude. Thus, applying TLS on a given set of variables could be an alternative approch to evaluate the goodness of a multipole relationship formed between the variables of the set. However, TLS does not provide any approach for searching through a large set of variables, e.g., the time series that capture temperature on the Earth's surface. Thus, it cannot not be used as a tool to find all the subsets of variables forming multipole patterns from a larger dataset. 


\textbf{Structure Learning:} Another stream of related work in machine learning literature is that of structure learning methods that learn the structure of stochastic dependencies among variables in a dataset in the form of a graphical model called \emph{Markov network}\cite{friedman2008sparse,meinshausen2006high},  which is a graph where each node represents a variable and follows the pairwise Markovian property, according to which it is independent of any non-neighboring node in the network conditioned on all of its neighboring nodes. Markov networks are typically studied to infer the conditional independence between different subsets of variables using various statistical inference techniques \cite{koller2009probabilistic}. In contrast the multipole patterns are defined to capture direct or indirect dependencies between different subsets of variables. An experimental demonstration on the limitations of structure learning methods in finding multipoles is provided in supplemental material.


\textbf{Correlation Networks:}  Linear relationships in time series data have also been studied in past using correlation networks, where each time series represent a node, and the weight of an edge between any two nodes represents the strength of the linear correlation between the corresponding time series. Correlation networks have been used in past for studying a variety of patterns, the most popular being `community`, which refers to a group of nodes (time series) with strong mutual positive correlations \cite{tsonis2011community,donges2009backbone}. If considered as a potential multipole, a community would have very low linear gain since its time series are highly similar, i.e., they show considerable collinearity. In contrast, time series in a multipole with high linear gain cannot be highly similar.

Some works, including our own \cite{kawale2011discovering}, have further studied pairs of negatively correlated communities, which form \emph{dipoles}. Multipoles  often have negative correlations among vectors---see Section \ref{sec:EmpObs}.  Nonetheless, those links can be relatively weak, i.e., not meaningful dipoles. Further, there is no guarantee that a dipole will show up as part of a multipole pattern. We also recently defined \emph{tripoles} \cite{agrawal2017tripoles}, but a multipole is not a generalization of a tripole. A tripole consists of a root and a pair of leaf time series, such that the sum of the leaf time series shows much stronger correlation with the root compared to either of their individual correlations with the root. Thus, as with regression, one of the variables (root or dependent variable) has a special role, which is not the case for multipoles. Further, tripoles are restricted to only one linear combination (i.e. sum of leaves), whereas multipoles allow arbitrary normalized linear combination to attain linear dependence. More importantly, there does not seem to be a way to generalize the tripole concept beyond three time series in a way that would facilitate efficient search for such patterns in a large data sets.

In summary, the problem of finding multipoles in the data is a novel and unique problem and to the best of our knowledge, there doesn't exist any method in the relevant literature that is directly suitable to solve this problem. 

\input{Methods/FindingMultipolesJan2019}

\section{Data and Experimental Evaluation}\label{Sec:Res}
In this section, we discuss results and computational evaluation of proposed approach CoMEtExtended. Specifically, we evaluate the completeness of CoMEtExtended against a regularized linear regression-based baseline, analyze trade-off between completeness and computational efficiency of CoMEtExtended at multiple parameter settings, study the scalability of CoMEtExtended, analyze the statistical significance of the multipoles, and evaluate their utility using real-world datasets from climate science and neuroscience domains. All experiments were run on a computer with 20 processors, each processor being Intel(R) Xeon(R) CPU E5-2470 0 running at 2.30GHz with a total shared RAM of 100 GB, running Linux version 2.6.32-696. We begin by describing all the datasets along with the data pre-processing steps that were applied to each of them. 
\subsection{Data and Preprocessing}\label{Sec:Preproc}
\subsubsection{Sea Level Pressure (SLP) data:} We used monthly Sea Level Pressure (SLP) dataset provided by NCEP/National Center for Atmospheric Research (NCAR) Reanalysis Project \cite{kistler2001ncep}, which is available from 1979-2014 (36 years) at a spatial resolution of 2.5 $\times$ 2.5 degree (10512 grid points, also referred to as locations). In this paper, we constructed SLP time series for each location using only the months of winter season (December, January, and February) from each year, thereby resulting in 108 observations in every time series. For each of the time series, we followed the standard pre-processing steps followed in climate science to remove the annual seasonality and linear trends \cite{kawale2013graph}. 

Relationships in climate datasets are preferably studied between regions (sets of spatially contiguous locations) as opposed to individual locations because of spatial autocorrelation, due to which locations in a spatial neighborhood have highly similar time series that will lead to discovery of redundant relationships. Therefore, we next converted the given location-based time series dataset into a set of 171 region-based time series dataset using a simple clustering procedure that is described in supplemental material. In addition, for purposes of validation of obtained multipoles, we used monthly Hadley Center SLP (HadSLP2) observational data available for years prior to 1979 to obtain the time series of these regions. 

\subsubsection{Brain fMRI data:} We used neuroimaging data collected at the University of Utah as part of a reproducibility study \cite{anderson2011reproducibility}. In this study, a set of 50 functional-Magnetic Resonance Imaging (fMRI) scans of one subject were acquired while the subject was involved in an audio-visual task (watching cartoons). Another set of 50 fMRI scans were collected from the same subject while the subject was resting. The spatial resolution and the temporal resolution of every scan was 3mm $\times$ 3mm $\times$  3mm and 2 secs, respectively. A number of fMRI pre-processing steps---described in \cite{anderson2011reproducibility}---were performed including motion correction, unwarping, and filtering. In addition, we used an Automated Anatomical Labeling Atlas \cite{tzourio2002automated}, which maps grey matter locations to 90 anatomical regions, to compute a mean time series of each brain region from each scan. As a result, we obtained a set of 90 time series for each of the 100 fMRI scans. We applied our approach to find multipoles in one of the 50 audio-visual fMRI scans, while the other 49 scans were used for evaluation purposes that we will describe later in this section.

\subsection{Parameter settings}
Two user-specified thresholds, minimum linear gain ($\delta$) and minimum linear dependence ($\sigma$), are needed for discovering multipoles. The choice of values for these parameters needs to be determined based on domain knowledge and the availability of computational resources to find multipoles. In particular, a relaxed linear gain threshold $\delta$ will increase the search space of multipoles and thus will require more computational time for search and evaluation. Similarly, a lower threshold of $\sigma$ will result in a larger number of discovered multipoles to be analyzed further by domain experts. 

In this work, we performed computational evaluation and scalability analysis at different combinations of values of $\sigma \in \{0.4,0.5,0.6\}$ and $\delta \in \{0.1,0.15,0.2\}$ respectively. Statistical significance analysis was performed on multipoles obtained at $\sigma= 0.50$ and $\delta= 0.15$ for both SLP and brain fMRI datasets.

\begin{table}[h]
\footnotesize
\centering
\centering
\begin{tabular}{|c|c|c|c|}
\hline 
&Total Multipoles in &\multicolumn{2}{|c|}{Completeness} \\
 \cline{3-4}
\textbf{($\sigma$,$\delta$)} & Pseudo-complete set &LAB &CoMEtExtended \\
\hline
\textbf{(0.4,0.1)} &70150 &0.09\% &81\%, $\rho$=$0.01$\\ \hline 
\textbf{(0.4,0.15)} &6255 &0.33\% &96\%, $\rho$=$0.01$\\ \hline 
\textbf{(0.4,0.2)} &1264 &0.39 \% &99\%, $\rho$=$0.01$\\ \hline 
\textbf{(0.5,0.1)} &41126 &0.15\% &76\%, $\rho$=$0.01$\\ \hline 
\textbf{(0.5,0.15)} &3348 &0.62\% &92\%, $\rho$=$0.01$\\ \hline 
\textbf{(0.5,0.2)} &930 &0.54\% &99\%, $\rho$=$0.01$\\ \hline 
\textbf{(0.6,0.1)} &13743 &0.47\% &75.5\%, $\rho$=$0.03$\\ \hline 
\textbf{(0.6,0.15)} &1525 &1.38\% &85\%, $\rho$=$0.01$\\ \hline 
\textbf{(0.6,0.2)} &488 &1.0\% &98\%, $\rho$=$0.01$\\ \hline 
\end{tabular}
\caption{Completeness evaluation of CoMEtExtended against LASSO-based baseline (LAB)  at different combinations of $\sigma$ and $\delta$ in the SLP dataset. The parameter $\rho$ in CoMEtExtended was set so as to keep the computational time under 90 minutes.}
\label{tab:compevalSLP}
\end{table}

\begin{table}[h]
\footnotesize
\centering
\centering
\begin{tabular}{|c|c|c|c|}
\hline 
&{\footnotesize Total Multipoles in} &\multicolumn{2}{|c|}{Completeness} \\
 \cline{3-4}
\textbf{($\sigma$,$\delta$)} & {\footnotesize Pseudo-complete set} &LAB &CoMEtExtended \\
\hline
\textbf{(0.4,0.1)} &15855 &0.006\% &71\%, $\rho$=$0.2$\\ \hline 
\textbf{(0.4,0.15)} &3019 &0\% &98\%, $\rho$=$0.2$\\ \hline 
\textbf{(0.4,0.2)} &716 &0\% &100\%, $\rho$=$0.2$ \\ \hline 
\textbf{(0.5,0.1)} &15258 &0.006\% &70\%, $\rho$=$0.2$\\ \hline 
\textbf{(0.5,0.15)} &2805 &0\% &98\%, $\rho$=$0.2$\\ \hline 
\textbf{(0.5,0.2)} &697 &0\% &100\%, $\rho$=$0.2$\\ \hline 
\textbf{(0.6,0.1)} &13721 &0.007\% &82\%, $\rho$=$0.25$\\ \hline 
\textbf{(0.6,0.15)} &2172 &0\% &97\%, $\rho$=$0.2$\\ \hline 
\textbf{(0.6,0.2)} &547 &0\% &100\%, $\rho$=$0.2$ \\ \hline 
\end{tabular}
\caption{Same as Table~\ref{tab:compevalSLP}, but in the fMRI dataset.}
\label{tab:compevalfMRI}
\end{table}

\subsection{Evaluation}\label{Sec:CompEval}
Comparison of CoMEtExtended to another approach is difficult, since it defines local patterns in terms of linear algebra concepts and finds such patterns by searching for negative cliques in a similarity graph. None of the related works mentioned in Section \ref{sec:RelWork} do the same thing. Nonetheless, to provide some comparison, we evaluate the completeness of the search of CoMEtExtended with respect to a LASSO-based baseline approach (LAB) that is described as follows:

\textbf{LASSO-based baseline approach (LAB):} LASSO is a variant of regularized linear regression that obtains a subset of variables from a larger set that could be linearly combined to predict the given predictand with high accuracy. Specifically, given a predictand $Y$ and a set of predictors $\mathbf{X} = [X_1,X_2,...,X_n]$, it learns a sparse set of regression coefficients $\mathbf{\beta} = [\beta_1,\beta_2,...,\beta_k]^T$ by minimizing the following objective function:
\begin{align*}
    \min\limits_{\mathbf{\beta}}||Y - \mathbf{X}\mathbf{\beta}||_2 + \lambda||\mathbf{\beta}||_1,
\end{align*}
where $\lambda$ is the hyperparameter that can be tuned to control the number of non-zero regression coefficients in the solution. In this baseline, we use LASSO to find potential candidates that could form multipoles. Specifically, in a dataset of $N$ variables $\{X_1,X_2,...,X_n\}$, for any variable $X_i$, we consider the remaining set of variables as predictors and apply LASSO find a subset of predictors from the remaining $N-1$ variables that could be linearly combined to best predict $X_i$. The combined set of $X_i$ and the predictors with non-zero regression coefficients is then added to the set of potential candidates of multipoles. For each variable $X_i$, we obtain up to $N-1$ different solutions of LASSO by varying  $\lambda$ such that every solution yields a different number of predictors with non-zero regression coefficients.  By applying the above procedure to all possible $X_i$, we obtain upto $N\times (N-1)$ potential candidates of multipoles in total. For each the obtained candidates, we next compute linear dependence and linear gain and discard all those that  do not satisfy the given thresholds $\sigma$ and $\delta$. 


We next evaluate the completeness of CoMEtExtended with that of LAB.The completeness of an algorithm is measured as the fraction of multipoles of a \emph{complete set} (a set that includes all the multipoles present in the data) that it finds from the data. An ideal approach to generate a complete set of multipoles will be to perform an exhaustive brute-force search over all possible subsets of variables and select those subsets that show linear dependence $\geq \sigma$ and linear gain $\geq \delta$. However, in practice, such an approach is computationally infeasible even for small-size datasets. For instance, in case of SLP data that has 171 time series, the estimated time on our computer to examine all subsets of size 5 is more than 5 days, size 6 is more than 144 days, and so on. Hence, for the purposes of evaluation, we generated a \emph{pseudo-complete} set of multipoles that includes unique and non-redundant multipoles generated by the following approaches: i) an exhaustive brute-force search over all subsets of variables upto size 4 for SLP, and 5 for fMRI dataset, ii) A random-search approach that is run for 24 hours to examine random subsets of variables of size $k$, where $k \geq 5 $ for SLP, and $k \geq 6$ for fMRI datasets, and select the ones that show linear dependence $\geq \sigma$ and linear gain $\geq \delta$, iii) LAB approach, and iv) CoMEtExtended approach, where the choice of parameters 
is set so as to keep its computational time within 90 minutes. 

Tables~\ref{tab:compevalSLP} and \ref{tab:compevalfMRI} summarize the results of completeness evaluation on SLP and fMRI datasets respectively. The first column in both the tables indicate the total size of the pseudo-complete set of multipoles at different combinations of parameters $\sigma$ and $\delta$. Second and third columns indicate the completeness of LAB and CoMEtExtended approach. As can be seen in both tables, LAB recovers a negligible or very tiny fraction of multipoles in all the parameter settings. For instance, in the SLP dataset, LAB recovers less than  1.3\% of total multipoles in pseudo-complete set in all the parameter settings. The completeness is negligible ($<0.1\%$) for cases where $\delta = 0.1$. Further, in the fMRI dataset, LAB is able to recover only one multipole for cases where $\delta = 0.1$, whereas for the remaining other cases where $\delta \geq 0.15$, LAB is unable to find any multipole. This strongly indicates that linear regression-based approaches like LASSO are not suitable for finding multipoles in the data. 

In contrast, the completeness of CoMEtExtended is more than 80 \% for all the parameter settings. Moreover, the completeness of CoMEtExtended increases at higher thresholds of linear gain and is close to 100 \% for $\delta = 0.2$, which shows that our approach is less likely to miss multipoles with high linear gain. This is in concordance with our empirical observations made in Figure~\ref{fig:EmpRes}, according to which, whenever linear gain is high, the largest pairwise correlation in the self-canceling version of all of the multipoles tends to be much lower and approaches strong negative values. As a result, they are more likely to be captured as promising candidates by our approach. In summary, CoMEtExtended outperforms the LASSO-based baseline approach in completeness at different thresholds on linear dependence and linear gain. Further, it is much more  efficient and relatively complete in finding multipoles at higher thresholds of linear gain, compared to brute-force and the LASSO-based baseline approach.


\textbf{Trade-off between completeness and efficiency:} We also evaluated the performance of CoMEtExtended based on the trade-off made between completeness and efficiency by varying parameter $\rho$. Table~\ref{tab:complvstimeSLP} shows completeness and computational time (in seconds) on SLP and fMRI datasets respectively at different values of $\rho$ for different combinations of $\sigma$ and $\delta$. As $\rho$ increases, the completeness improves but also adds to the total computational cost. This is expected since $\rho$ signifies the upper bound on the largest pairwise correlation in the self-canceling version of a promising candidate. Therefore, as $\rho$ increases, more sets qualify as promising candidates, which further expands the search space, leading to higher computational cost.

\begin{table}[h]
\small
\centering
\begin{tabular}{|c|c|c|c|c|}
\hline 
&\multicolumn{4}{|c|}{Completeness, Computing Time (in minutes)} \\
 \cline{2-5}
 \textbf{($\sigma$,$\delta$)} &$\rho=-0.2$ &$\rho=-0.1$ &$\rho=0$ &$\rho=0.01$ \\
\hline
\textbf{(0.4,0.1)} &1\%,0.1 &12\%,0.5 &77\%,13 &81\%,28 \\ \hline 
\textbf{(0.4,0.15)} &21\%,0.1 &76\%,0.3 &95\%,9 &96\%,21 \\ \hline 
\textbf{(0.4,0.2)} &73\%,0.1 &92\%,0.2 &99\%,9 &99\%,20 \\ \hline 
\textbf{(0.5,0.1)} &2\%,0.1 &12\%,0.3 &71\%,11 &76\%,23 \\ \hline 
\textbf{(0.5,0.15)} &27\%,0.1 &63\%,0.2 &91\%,8 &72\%,18 \\ \hline 
\textbf{(0.5,0.2)} &70\%,0.1 &90\%,0.2 &98\%,7 &99\%,17 \\ \hline 
\textbf{(0.6,0.1)} &6\%,0.1 &14\%,0.2 &63\%,7 &68\%,14 \\ \hline 
\textbf{(0.6,0.15)} &31\%,0.1 &55\%,0.2 &82\%,6 &85\%,12 \\ \hline 
\textbf{(0.6,0.2)} &67\%,0.1 &83\%,0.2 &97\%,5 &98\%,11 \\ \hline
\end{tabular}
\caption{Performance of CoMEtExtended at different values of $\rho$ for different combinations of $\sigma$ and $\delta$ in SLP dataset. Each cell in the table contains two values: i)Completeness of search, and ii) Computational time (in minutes) taken by CoMEtExtended.}
\label{tab:complvstimeSLP}
\end{table}

Also note that at higher $\delta$, the completeness of CoMEtExtended reaches close to 100 \% at much smaller values of $\rho$ and requires much less computing time. This indicates that our approach is much more efficient in finding multipoles with high linear gain. This is again consistent with the empirical observations made in Figure~\ref{fig:EmpRes} where we observed that for a multipole with high linear gain, all the pairwise correlations in its self-canceling version tend to have stronger negative values. Therefore, they get included among the promising candidates at much lower values of $\rho$. 

\subsection{Scalability Analysis} \label{sec:scale}
As is common with many pattern finding techniques, such as frequent pattern mining, the CoMEt algorithm is inherently exponential. (For a detailed analysis of the the time complexity  of the the three parts of the algorithm, see the Supplemental material.)  However, as with association analysis, adjusting the parameter settings---in this case, $\sigma_S$ and $\Delta\sigma_S$---can make the pattern search quite tractable in many situations. 
To illustrate, we next discuss the scalability of our approach using SLP data, i.e., how does the computational time vary with the size of datasets (number of time series). To obtain datasets of different sizes, we first generated 10 additional seasonal datasets for different seasons, each season being a set of three consecutive months: ((Jan.,Feb.,Mar.), (Feb.,Mar.,Apr.) ,...(Oct. Nov. Dec.)). Each dataset consists of time series from the same 171 regions that we chose for our original SLP dataset. We then generated 10 datasets of sizes 171*k, where $k\in [1,10]$ by merging k of the above 10 seasonal datasets.  

\begin{figure}
    \centering
    \begin{subfigure}[t]{0.20\textwidth}
    \includegraphics[width=\textwidth]{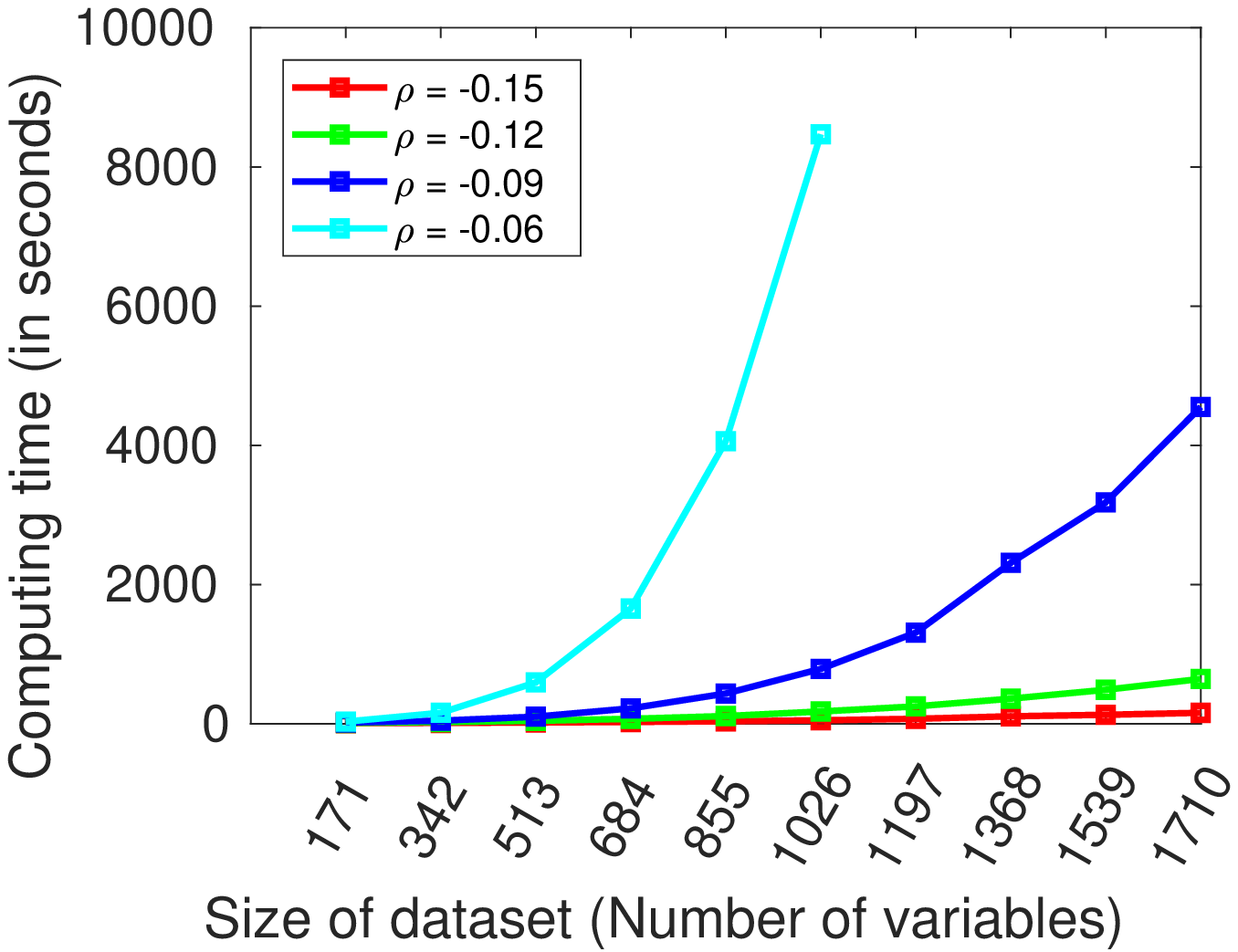}

    \centering
    \caption{$\sigma$ = 0.5, $\delta$= 0.15}
    \label{fig:RunningTime1}
    \end{subfigure}
    \centering
    \begin{subfigure}[t]{0.20\textwidth}
    \includegraphics[width=\textwidth]{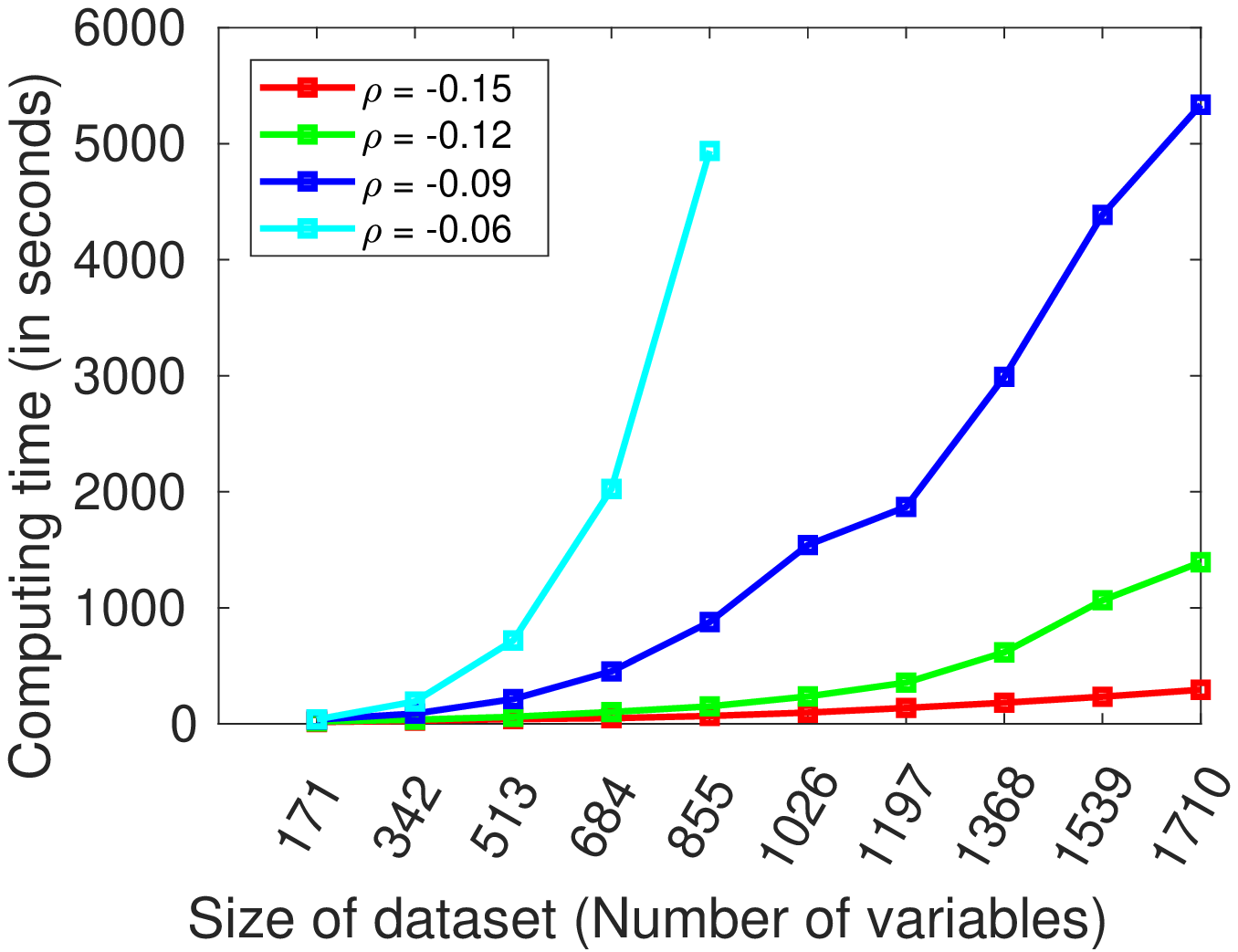}
    \centering
    \caption{$\sigma$ = 0.4, $\delta$= 0.15}
    \label{fig:RunningTime2}
    \end{subfigure}
     \begin{subfigure}[t]{0.20\textwidth}
    \includegraphics[width=\textwidth]{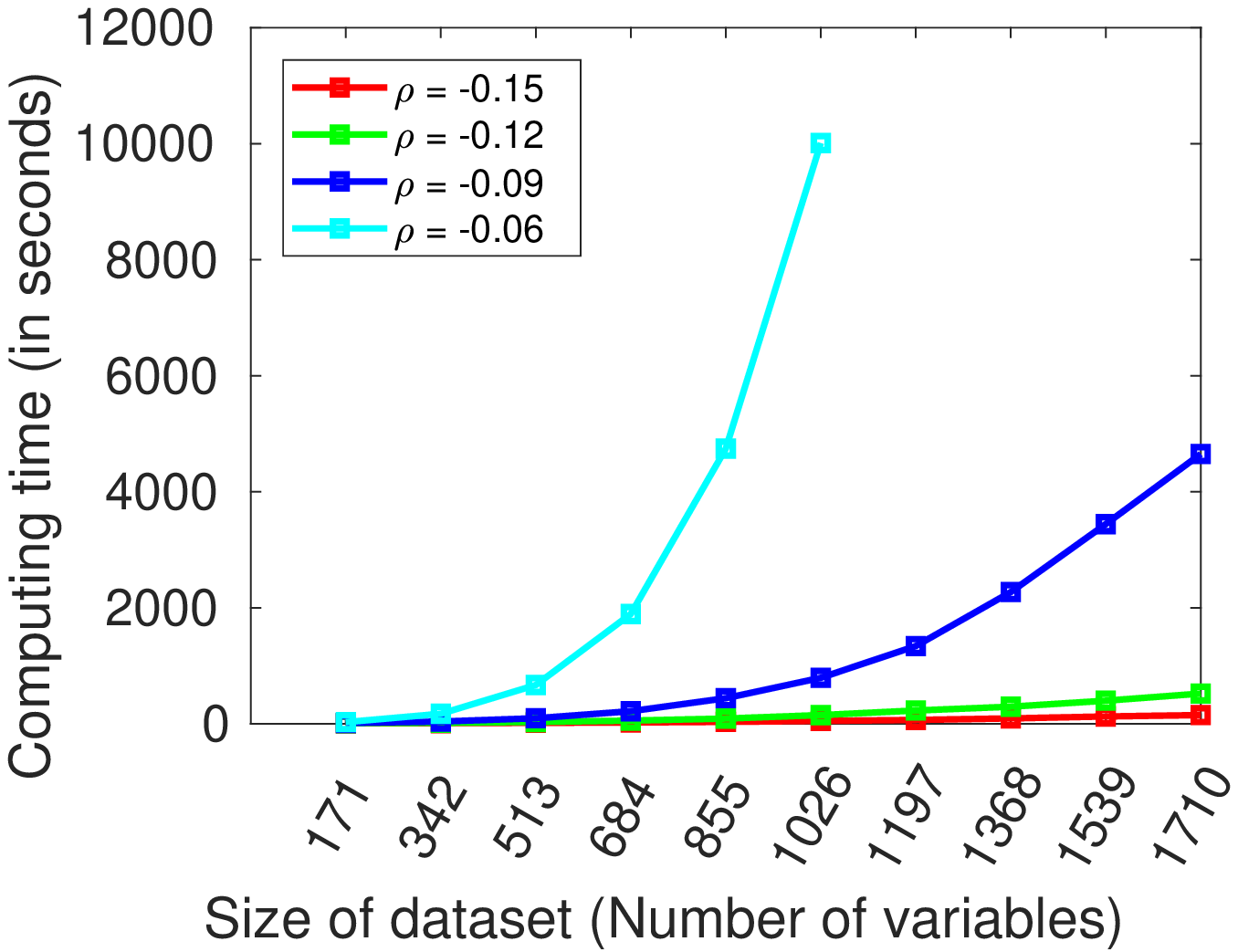}
    \centering
    \caption{$\sigma$ = 0.4, $\delta$= 0.2}
    \label{fig:RunningTime3}
    \end{subfigure}
     \begin{subfigure}[t]{0.20\textwidth}
    \includegraphics[width=4cm,height=3cm]{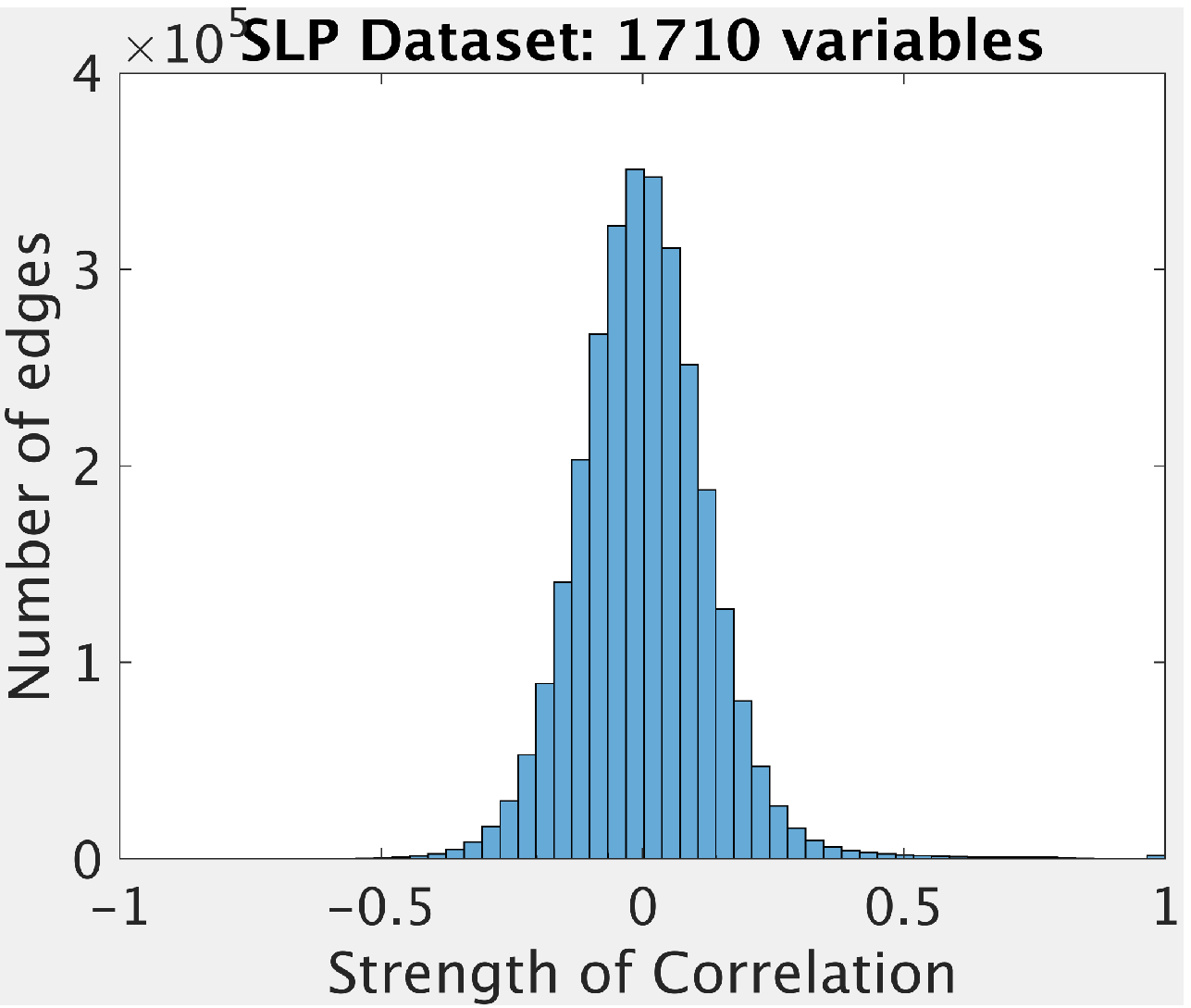}
    \centering
    \caption{Distribution of correlation strengths}
    \label{fig:CorrHist}
    \end{subfigure}
\caption{\textbf{Scalability analysis:} Figures~\ref{fig:RunningTime1}, \ref{fig:RunningTime2}, and \ref{fig:RunningTime3} plot the computing time (Y-axis) of COMETExt on different sizes of SLP datasets (X-axis) at different values of $\rho$ for three parameter settings (indicated in subcaptions). Figure~\ref{fig:CorrHist} shows the distribution of correlation strengths of edges in a correlation network in the SLP dataset that has 1710 time series. See section~\ref{sec:scale} for further details.}
\label{fig:Scalability}
\end{figure}

Figure~\ref{fig:RunningTime1} shows the total computational time of CoMEtExtended on all of the above datasets, at $\sigma =0.5$ and $\delta = 0.15$ for different values of $\rho$ in range $[-0.15,-0.06]$. As can be seen in figure, the computing time increases with the increase in the size of the datasets at different rates depending on $\rho$. For stronger negative values of $\rho$, the computing time increases almost linearly with the increase in size of datasets, which highlights its scalability\footnote{We have also demonstrated the scalability of our approach on larger synthetic datasets containing upto 100k time series (see supplemental)}. However, as $\rho$ approaches zero, the scalability is weak, and the computing time increases dramatically for bigger datasets. Similar observations are also made at other parameter settings, e.g. at $(\sigma = 0.4, \delta = 0.15)$ and $(\sigma = 0.4, \delta = 0.2)$, as shown in Figures~\ref{fig:RunningTime2} and \ref{fig:RunningTime3} respectively. The observed loss in scalability could be attributed to the typical distribution of pairwise correlations in the correlation graph of any time series dataset, as shown in Figure~\ref{fig:CorrHist} for one of the SLP datasets. The distribution is bell-shaped with major fraction of edges having strengths close to zero in magnitude. Consequently, as $\rho$ approaches zero, the number of cliques found in Step 1 of the algorithm increase exponentially. Note that many of the cliques that have all or most of the edges being weak are expected to have weak linear dependence among their variables, and hence are unlikely to form a multipole. By setting $\rho$ to stronger negative values, we avoid such cliques and save a lot of computing time. However, that also leads to missing some of the interesting cliques where only one or two edges were close to zero.\footnote{The exact number of missing cliques could not be computed due to the absence of ground truth and computational intractability of brute-force approach.} Such cliques could potentially be recovered by heuristic approaches, which could be an interesting direction to pursue for future work.

\subsection{Evaluation of Multipoles}\label{Sec:Eval}
One of the key challenges of this work is distinguishing between reliable and spurious multipoles, i.e., those multipole patterns that arise due to random variation, from the large number of discovered multipoles. Domain validation is an ideal approach to evaluate multipoles, but most of the multipole relationships discovered in this work are currently unknown to domain scientists. Thus, in our work, we used an empirical evaluation framework that consists of two steps. The first step involves a procedure for estimating the statistical significance of a multipole. This procedure is then used in the second step to assess the reproducibility of a discovered multipole in multiple datasets, where each dataset is collected during a time period different from that of original dataset used for finding multipoles. Intuitively, spurious multipoles are less likely to reproduce in  time periods that were not used for finding them.  In contrast, multipoles that do reproduce with high statistical significance are more likely to be  patterns that are outcome of a real phenomenon, and thus they would be ideal candidates for further investigation by domain experts.  

\subsubsection{Step 1: Statistical Significance Evaluation:}\label{sec:stateval}
To filter spurious multipoles, it is important to answer the following two questions: i) how likely is it that the observed level of linear dependence, $\sigma_S$, of a multipole $S$ is due to chance? and ii) does every member in $S$ contribute significantly to the linear dependence of $S$?

To address the first question, we generate a null distribution of linear dependence by randomly generating 100,000 sets of time series and evaluating each set $S$ for its level of linear dependence.  Each of these randomly generated sets is created by sampling time series from different time periods. For instance, for our SLP investigations, a random set of size $|S|$ is constructed by sampling a time series from any $|S|$ of the nine time windows of HadSLP2 data. Similarly, for brain fMRI data, a random set of $|S|$ time series is constructed by sampling a time series from any $|S|$ of the 50 scans. Generating a random set in this manner is an approximation to independently generating $|S|$ time series while ensuring that the general underlying nature of domain time series (e.g. autocorrelation, periodicity etc.) is retained in the randomly generated data. Using the resultant null distribution, we then determine the statistical significance of the multipoles we originally found. We evaluate  $\sigma_S$ at a 0.01 level of significance. 

We next describe our approach to assess the second point, i.e, the significance of the contribution of each of the $k$ variables in a given set $S = \{X_1,X_2,...,X_k\}$ to its linear dependence. Specifically, to assess the significance of the contribution of time series $X_i$, we replace it with a random time series $X_R$ that is sampled from an independent dataset in a manner similar to the procedure above. We then compute the linear dependence of the resultant set, which we call $S'$. If the contribution of $X_i$ is not spurious, it would be unlikely for a randomly chosen time series $X_R$ to replicate it in which case, $\sigma_{S'}\leq \sigma_S$. We repeat the above process 1000 times and compute the fraction of the population for which $\sigma_{S'}\leq \sigma_S$ holds true. This fraction is our significance level. We again use a significance level of 0.01. The above procedure is repeated for each of the $k$ members.

\begin{figure}
    \centering
    \begin{subfigure}[t]{0.20\textwidth}
    \includegraphics[width = 4.5cm,height = 3cm]{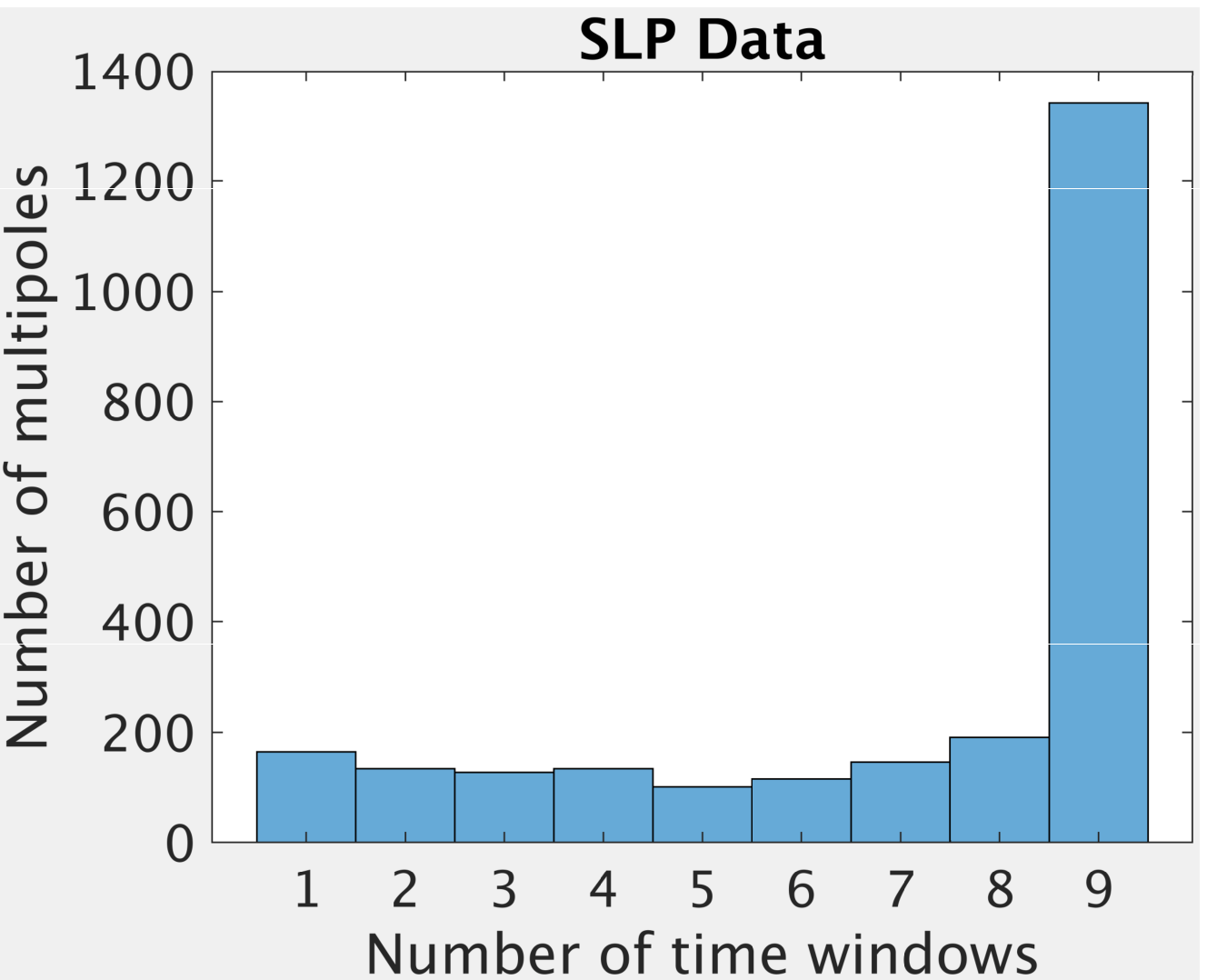}

    \centering
    \caption{SLP Data}
    \label{fig:Reprod_SLP}
    \end{subfigure}
    \centering
    \begin{subfigure}[t]{0.20\textwidth}
    \includegraphics[width = 4.5cm,height = 3cm]{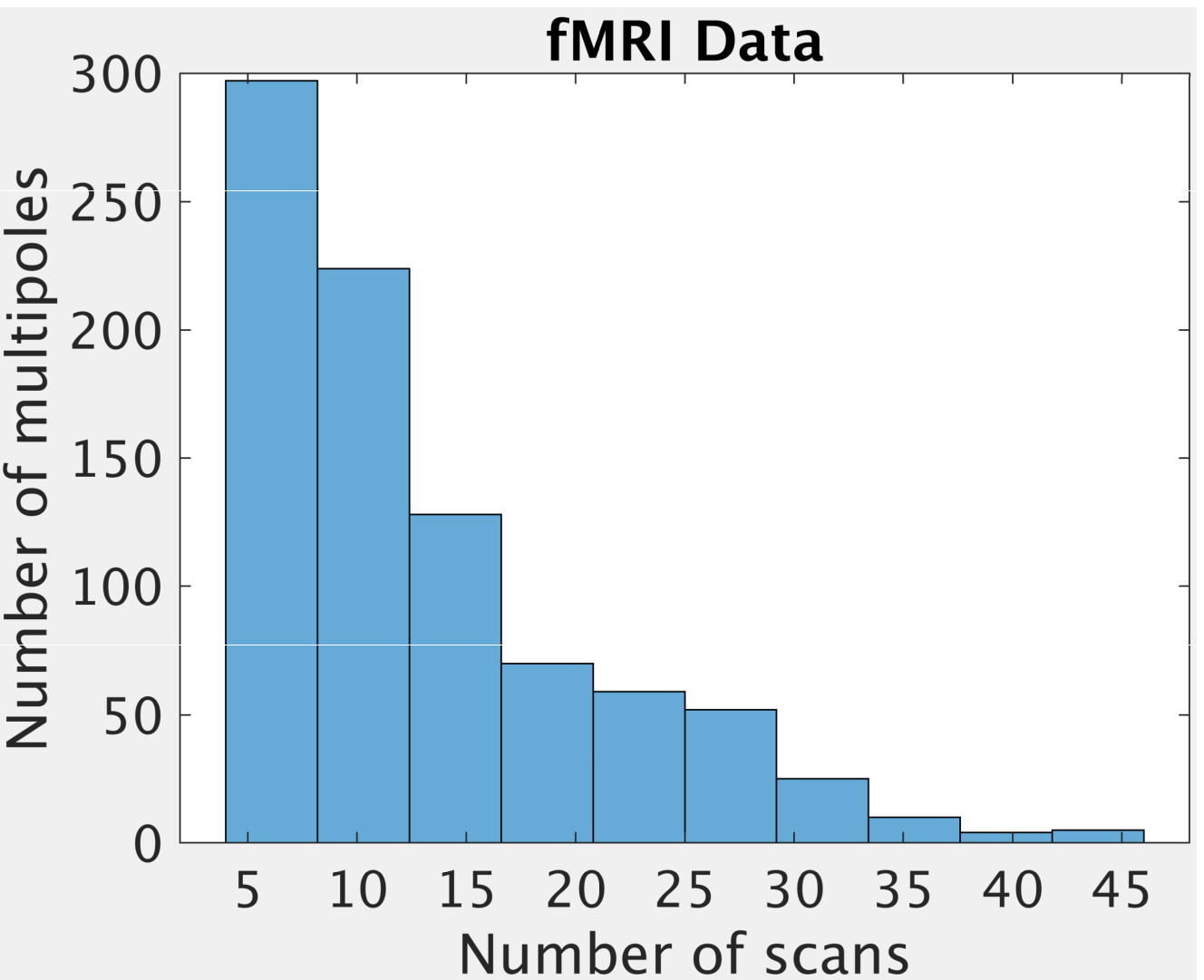}
    \centering
    \caption{fMRI Data}
    \label{fig:Reprod_fMRI}
    \end{subfigure}
\caption{\textbf{Reproducibility analysis:} Figures show the number of multipoles that are found to be reproducible (Y-axis) in different number of independent datasets (X-axis) for both SLP and fMRI data for $\sigma = 0.5$, $\delta = 0.15$, and $
\rho = 0$. See section~\ref{sec:reprod} for further details}
\label{fig:ReprodEval}
\end{figure}

\subsubsection{Step 2: Reproducibility in Independent Datasets}\label{sec:reprod}
In this step, we estimate the reproducibility of a given multipole in multiple time periods. A multipole is considered to be reproducible in a dataset $D$, if, at a 0.01 level of significance, it is found to have statistically significant linear dependence, as well as a statistically significant contribution from each of its members. Specifically, for each multipole discovered in 1979-2014 SLP dataset, we computed its linear gain and linear dependence in HadSLP2 data in 9 time windows 1901-1936, 1906-1941,...,1941-1976. Likewise, for multipoles discovered in one of the 50 brain fMRI scans, we studied their reproducibility in the remaining 49 scans.


Figure~\ref{fig:Reprod_SLP} shows number of multipoles reproduced in different numbers of HadSLP2 time windows during 1901-1976. More than 40\% of multipoles reproduced in all 9 time windows. Similarly, Figure~\ref{fig:Reprod_fMRI}  shows number of multipoles from fMRI data reproduced in different number of fMRI scans taken while the subject was watching a video. At least 25\% of multipoles reproduced in more than 10 other scans. Higher reproducibility of multipoles suggests that they are more robust to noise in the data and unlikely to be spurious.

\section{Case Studies}\label{Sec:PhysInt}

Results discussed in the previous section indicate the existence of several multipole relationships with high reproducibility in multiple independent time periods, which makes a compelling case for their connection to underlying physical phenomena that might be currently unknown to domain scientists, but could potentially be discovered by domain experts upon further analysis. In this section, we present case studies on the physical interpretation of two of the discovered multipoles in SLP and brain fMRI data.

\subsection{Discovering Climate Phenomena}
\begin{figure}
    \centering
    \includegraphics[scale=0.5]{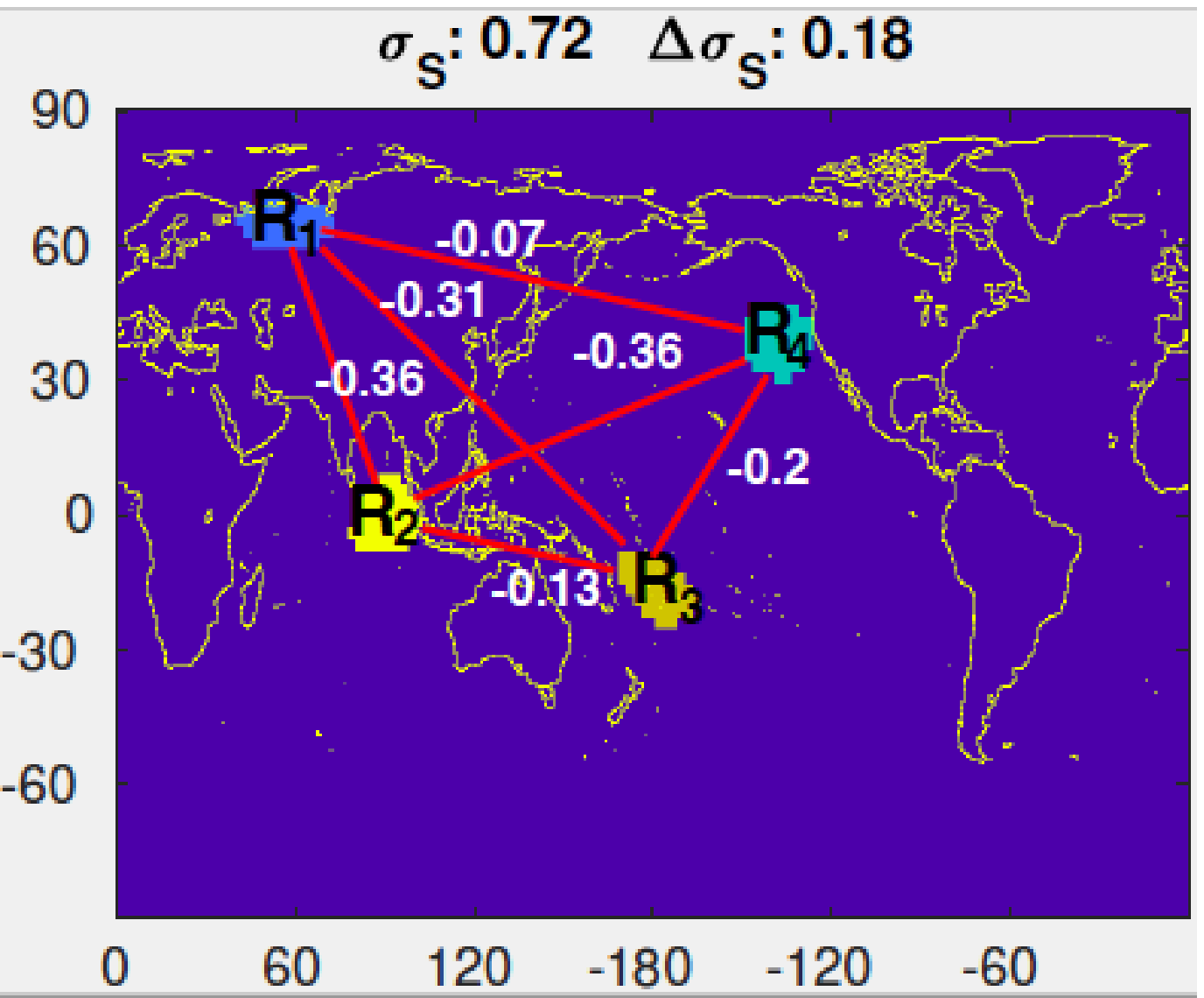}
    \caption{A 4-pole in 1979-2014 that was reproducible in 7 out of 9 time windows during 1901-1976 }
    \label{fig:CaseStudyClim}
\end{figure}

\begin{figure*}
    \centering
    \begin{subfigure}[t]{0.3\textwidth}
    \includegraphics[width=\textwidth]{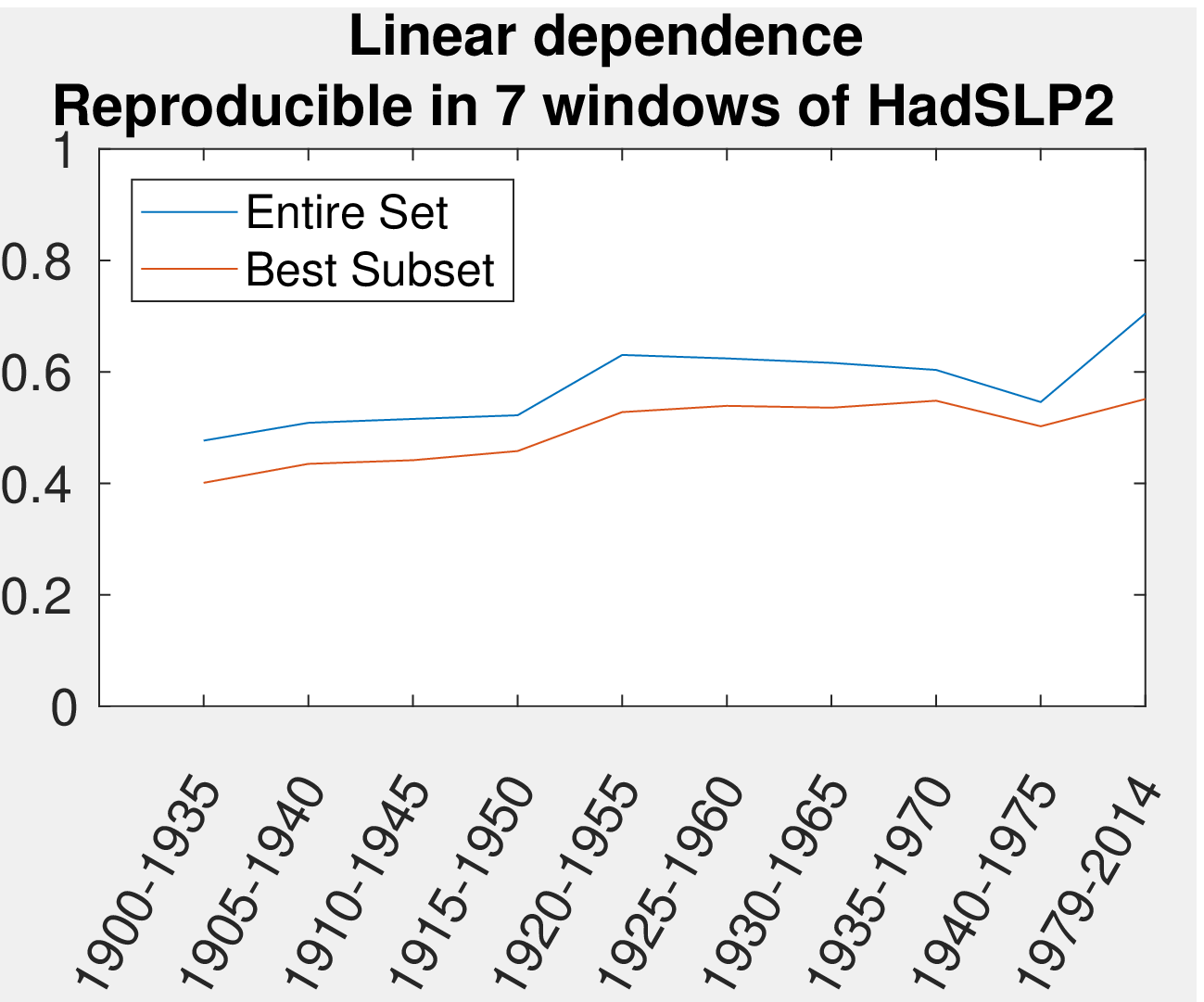}
    \centering
    \caption{HadSLP2}
    \label{fig:Reprod_HadSLP2}
    \end{subfigure}
    \centering
    \begin{subfigure}[t]{0.3\textwidth}
    \includegraphics[width=\textwidth]{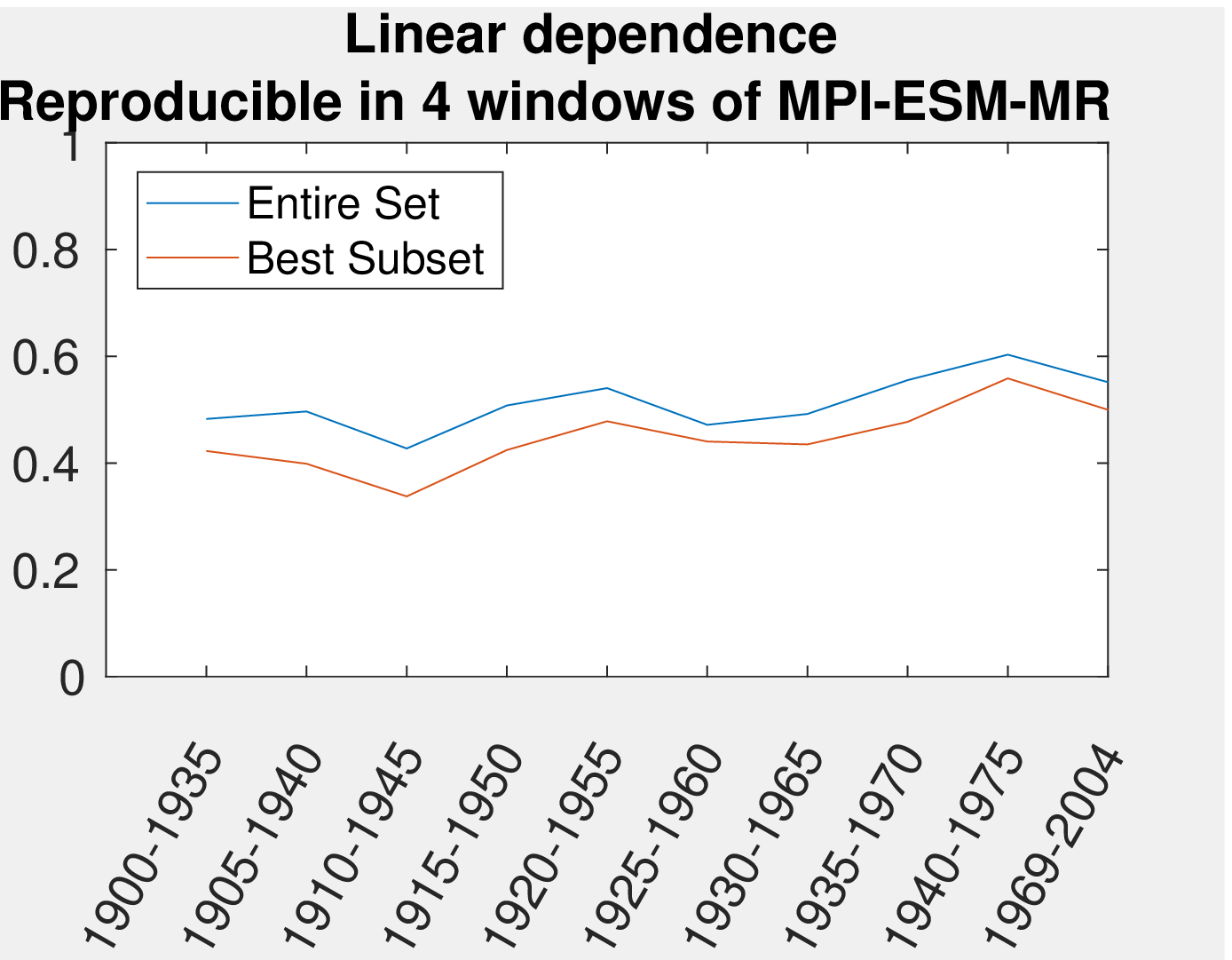}
    \centering
    \caption{Climate Model: MPI-ESM-MR}
    \label{fig:Reprod_MPI-ESM-MR}
    \end{subfigure}
    \centering
    \begin{subfigure}[t]{0.3\textwidth}
    \includegraphics[width=\textwidth]{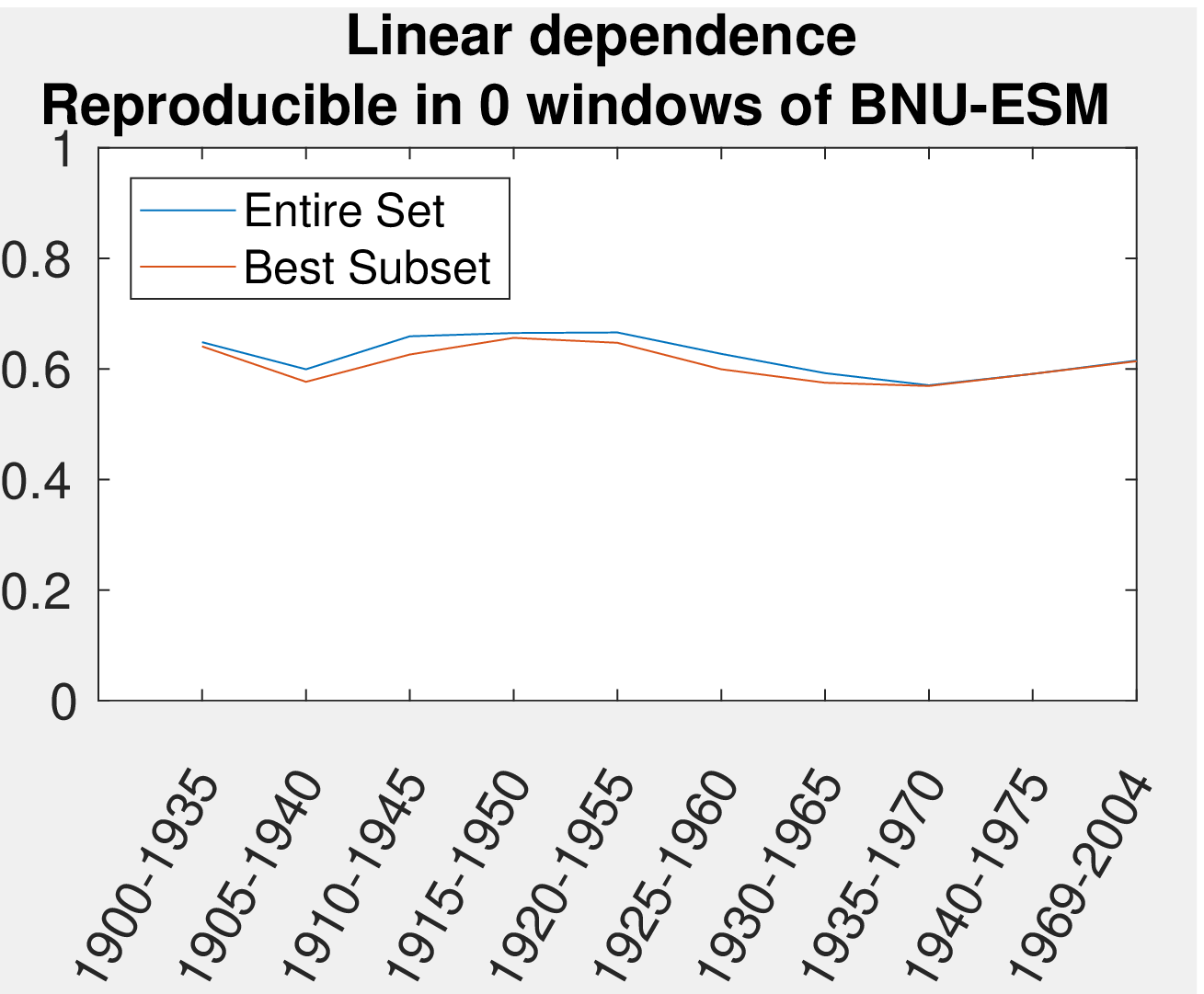}
    \centering
    \caption{Climate Model: BNU-ESM}
    \label{fig:Reprod_BNU-ESM}
    \end{subfigure}
\caption{\textbf{Climate Models Inter-comparison based on multipole simulations:} Summary statistics of multipole shown in Figure~\ref{fig:CaseStudyClim} in different time windows of HadSLP2 data (Figure~\ref{fig:Reprod_HadSLP2}) and different CMIP5 climate model datasets (Figures~\ref{fig:Reprod_MPI-ESM-MR} and \ref{fig:Reprod_BNU-ESM}). In each plot, the red curve indicates the strength of the linear dependence of the multipole, while the blue curve indicates the highest strength of linear dependence obtained for one of its subsets. The difference in red and blue curves thus indicate the linear gain of the multipole.}
\label{fig:ModelEval}
\end{figure*}

One of the multipoles in SLP data was found between the four regions shown in Figure~\ref{fig:CaseStudyClim}. The time series of the four regions show negative correlations with each other, resulting in a multipole relationship with linear dependence of $0.7$ and a linear gain of $0.15$ during 1979-2014. Further, as indicated in Figure~\ref{fig:Reprod_HadSLP2}, the multipole was found to be reproducible in 7 out of 9 time windows during the period of 1901-1976, showing strong linear dependence (red curve) and linear gain (indicated by gap between red and black curves). This multipole appears to be strongly related to the well-known climate phenomenon known as the El-Nino Southern Oscillation (ENSO) as indicated by regions $R_2$ and $R_3$. This phenomenon appears not only in the tropical Pacific Ocean, but also has large-scale impacts on regional climate outside the tropics \cite{vecchi2010nino}. More recently, a connection between the West Siberian Plain and ENSO was discovered as a \textit{tripole} relationship (as defined in \cite{agrawal2017tripoles}) between three regions, that are re-captured as regions $R_1$, $R_2$, and $R_3$ in above multipole. This phenomenon was attributed to a wave train that originates from the sub-tropical Atlantic and propagates north-eastwards towards the north of the West Siberian Plain, where it is deflected southeastwards and reaches all the way to the central Pacific Ocean, where the two centres of action of ENSO are located. Notably, region $R_4$ in the northern Atlantic Ocean is located near the proposed location of origin of the wave train. This finding can be further used to study the detailed path of the wave train and to attribute weather and climate characteristics over wave-affected regions to their potential ENSO origins.

\textbf{Evaluation of Climate Models:} Multipoles, being potential representatives of physical processes, could serve as useful benchmarks to evaluate various climate models that are often used to study climate change under different greenhouse gas emission scenarios. In particular, climate models can be evaluated based on their ability to reproduce the physical processes represented by these multipoles. For instance, Figure~\ref{fig:ModelEval} compares the linear dependence and linear gain of the above multipole across multiple time windows obtained in observations data (HadSLP2) and a  couple of climate models used in the IPCC (Intergovernmental Panel on Climate Change) CMIP5 (Couple Model Intercomparison Project Phase 5) evaluation. It can be seen that climate models differ in their ability to simulate the multipole effectively. Specifically, climate model MPI-ESM-MR is able to reproduce multipole with statistically significance in at least 4 time windows, whereas the other model BNU-ESM could not reproduce the given multipole at all.

\subsection{Studying Complex Dynamics in Brain}
The notion of multipoles proposed in this paper is highly suited for capturing complex signaling relationships in the brain. For example, one of the multipoles we discovered in one of the brain fMRI scans captures a relationship between three brain regions: \textit{Right Frontal Inferior} ($T_2$), \textit{Right Parietal Inferior}($T_1$), and \textit{Right Temporal Pole Superior}($T_3$). This multipole was found to be reproducible in 23 out of 50 `task` scans (collected while the subject was watching a cartoon video) while only in 5 of the 50 other `resting state` scans (collected while subject was resting). This relationship is interesting for multiple reasons. First, the parietal and temporal regions are known to be the first recipients of the video and audio stimuli \cite{tachibana2011parietal}, respectively, that are presented to the subject, in our case in the form of cartoons. Second, the dominance of the above multipole relationship in task scans in comparison to the resting scans provides strong evidence for signaling from parietal and temporal regions to the frontal region, that appears to be triggered due to visual and auditory stimuli. While some existing studies \cite{saygin2008retinotopy} have observed activity in the frontal region due to visual and auditory stimuli, no direct signaling pathway between visual and auditor cortices with the frontal region is established. In summary, the multipole framework could serve as a promising tool for discovering previously unknown signaling pathways that exist in the brain.

\section{Conclusion and Future Work}\label{Sec:Conc}
In this paper, we introduced and formally studied a novel class of multivariate linear relationships called \emph{multipoles} in time series data. A multipole corresponds to a set of time series that show much stronger linear dependence compared to any of its subsets. We presented a series of empirical observations to show that most interesting multipoles could be found as cliques of negative correlations in a correlation network, and proposed a novel and computationally efficient correlation network-based approach to find multipoles in the data. We demonstrated the utility of our proposed approach to find multipoles on real-world datasets from climate and neuroscience domain. Furthermore, we presented case studies from both domains to highlight the potential of multipoles in discovering novel physical processes. 
While the approach proposed in this paper is based on a series of empirical observations, it is noteworthy that all of our observations are universal in nature as opposed to being specific to a particular time series dataset. Moreover, there are certain scenarios in which our approach could be empirically shown to guarantee completeness of the search (see supplemental material for further details). Derivation of theoretical proofs of these observations are subject of future research. Other useful extensions of this work could be to extend the notion of multipoles to non-linear relationships, and generalization to time-lagged mutipole-relationships.

\ifCLASSOPTIONcompsoc
  \section*{Acknowledgments}
\else
  \section*{Acknowledgment}
\fi

We would like to thank Siddhant Agrawal, Department of Mathematics, University of Michigan for invaluable discussions. This work was supported by NSF grants IIS-1029771 and IIS-1319749 and NASA grant 14-CMAC14-0010. Access to the computing facilities was provided by the University of Minnesota Supercomputing Institute.
\ifCLASSOPTIONcaptionsoff
  \newpage
\fi

\let\OLDthebibliography\thebibliography
\renewcommand\thebibliography[1]{
  \OLDthebibliography{#1}
  \setlength{\parskip}{0pt}
  \setlength{\itemsep}{0pt plus 0.3ex}
}

\footnotesize
\bibliographystyle{IEEEtran}

\begin{thebibliography}{10}
\providecommand{\url}[1]{#1}
\csname url@samestyle\endcsname
\providecommand{\newblock}{\relax}
\providecommand{\bibinfo}[2]{#2}
\providecommand{\BIBentrySTDinterwordspacing}{\spaceskip=0pt\relax}
\providecommand{\BIBentryALTinterwordstretchfactor}{4}
\providecommand{\BIBentryALTinterwordspacing}{\spaceskip=\fontdimen2\font plus
\BIBentryALTinterwordstretchfactor\fontdimen3\font minus
  \fontdimen4\font\relax}
\providecommand{\BIBforeignlanguage}[2]{{%
\expandafter\ifx\csname l@#1\endcsname\relax
\typeout{** WARNING: IEEEtran.bst: No hyphenation pattern has been}%
\typeout{** loaded for the language `#1'. Using the pattern for}%
\typeout{** the default language instead.}%
\else
\language=\csname l@#1\endcsname
\fi
#2}}
\providecommand{\BIBdecl}{\relax}
\BIBdecl

\bibitem{kawale2013graph}
J.~Kawale, S.~Liess, A.~Kumar, M.~Steinbach, P.~Snyder, V.~Kumar, A.~R.
  Ganguly, N.~F. Samatova, and F.~Semazzi, ``A graph-based approach to find
  teleconnections in climate data,'' \emph{Statistical Analysis and Data
  Mining: The ASA Data Science Journal}, vol.~6, no.~3, pp. 158--179, 2013.

\bibitem{taschetto2014cold}
A.~S. Taschetto, A.~S. Gupta, N.~C. Jourdain, A.~Santoso, C.~C. Ummenhofer, and
  M.~H. England, ``Cold tongue and warm pool {ENSO} events in {CMIP5}: mean
  state and future projections,'' \emph{Journal of Climate}, vol.~27, no.~8,
  pp. 2861--2885, 2014.

\bibitem{wallace1981teleconnections}
J.~M. Wallace and D.~S. Gutzler, ``Teleconnections in the geopotential height
  field during the northern hemisphere winter,'' \emph{Monthly Weather Review},
  vol. 109, no.~4, pp. 784--812, 1981.

\bibitem{atluri2016brain}
G.~Atluri, A.~MacDonald~III, K.~O. Lim, and V.~Kumar, ``The brain-network
  paradigm: Using functional imaging data to study how the brain works,''
  \emph{Computer}, vol.~49, no.~10, pp. 65--71, 2016.

\bibitem{tibshirani1996regression}
R.~Tibshirani, ``Regression shrinkage and selection via the lasso,''
  \emph{Journal of the Royal Statistical Society. Series B (Methodological)},
  pp. 267--288, 1996.

\bibitem{lozano2009spatial}
A.~C. Lozano \emph{et~al.}, ``Spatial-temporal causal modeling for climate
  change attribution,'' in \emph{Proceedings of the 15th ACM SIGKDD
  international conference on data mining}.\hskip 1em plus 0.5em minus
  0.4em\relax ACM, 2009, pp. 587--596.

\bibitem{chatterjee2012sparse}
S.~Chatterjee, K.~Steinhaeuser, A.~Banerjee, S.~Chatterjee, and A.~R. Ganguly,
  ``Sparse group lasso: Consistency and climate applications.'' in
  \emph{SDM}.\hskip 1em plus 0.5em minus 0.4em\relax SIAM, 2012, pp. 47--58.

\bibitem{carroll2006measurement}
R.~J. Carroll, D.~Ruppert, C.~M. Crainiceanu, and L.~A. Stefanski,
  \emph{Measurement error in nonlinear models: a modern perspective}.\hskip 1em
  plus 0.5em minus 0.4em\relax Chapman and Hall/CRC, 2006.

\bibitem{barnston1987classification}
A.~G. Barnston and R.~E. Livezey, ``Classification, seasonality and persistence
  of low-frequency atmospheric circulation patterns,'' \emph{Monthly weather
  review}, vol. 115, no.~6, pp. 1083--1126, 1987.

\bibitem{ding2005circumglobal}
Q.~Ding and B.~Wang, ``Circumglobal teleconnection in the northern hemisphere
  summer,'' \emph{Journal of Climate}, vol.~18, no.~17, pp. 3483--3505, 2005.

\bibitem{friedman2008sparse}
J.~Friedman, T.~Hastie, and R.~Tibshirani, ``Sparse inverse covariance
  estimation with the graphical lasso,'' \emph{Biostatistics}, vol.~9, no.~3,
  pp. 432--441, 2008.

\bibitem{meinshausen2006high}
N.~Meinshausen and P.~B{\"u}hlmann, ``High-dimensional graphs and variable
  selection with the lasso,'' \emph{The annals of statistics}, pp. 1436--1462,
  2006.

\bibitem{tsonis2011community}
A.~A. Tsonis, G.~Wang \emph{et~al.}, ``Community structure and dynamics in
  climate networks,'' \emph{Climate dynamics}, vol.~37, no. 5-6, pp. 933--940,
  2011.

\bibitem{donges2009backbone}
J.~F. Donges, Y.~Zou, N.~Marwan, and J.~Kurths, ``The backbone of the climate
  network,'' \emph{EPL (Europhysics Letters)}, vol.~87, no.~4, p. 48007, 2009.

\bibitem{kawale2011discovering}
J.~Kawale, M.~Steinbach, and V.~Kumar, ``Discovering dynamic dipoles in climate
  data.'' in \emph{SDM}.\hskip 1em plus 0.5em minus 0.4em\relax SIAM, 2011, pp.
  107--118.

\bibitem{agrawal2017tripoles}
S.~Agrawal, G.~Atluri \emph{et~al.}, ``Tripoles: A new class of relationships
  in time series data,'' in \emph{Proceedings of the 23rd ACM SIGKDD
  International Conference on Data Mining}.\hskip 1em plus 0.5em minus
  0.4em\relax ACM, 2017, pp. 697--706.

\bibitem{van2004functional}
V.~G. van~de Ven, E.~Formisano, D.~Prvulovic, C.~H. Roeder, and D.~E. Linden,
  ``Functional connectivity as revealed by spatial independent component
  analysis of fmri measurements during rest,'' \emph{Human brain mapping},
  vol.~22, no.~3, pp. 165--178, 2004.

\bibitem{koller2009probabilistic}
D.~Koller and N.~Friedman, \emph{Probabilistic graphical models: principles and
  techniques}.\hskip 1em plus 0.5em minus 0.4em\relax MIT press, 2009.

\bibitem{eppstein2011listing}
D.~Eppstein and D.~Strash, ``Listing all maximal cliques in large sparse
  real-world graphs,'' in \emph{International Symposium on Experimental
  Algorithms}.\hskip 1em plus 0.5em minus 0.4em\relax Springer, 2011, pp.
  364--375.

\bibitem{cliquecode}
\url{https://github.com/darrenstrash/quick-cliques}.

\bibitem{kistler2001ncep}
R.~Kistler, W.~Collins, S.~Saha, G.~White, J.~Woollen, E.~Kalnay \emph{et~al.},
  ``The {NCEP}-{NCAR} 50-year reanalysis: Monthly means {CD-ROM} and
  documentation,'' \emph{Bulletin of the American Meteorological society},
  vol.~82, no.~2, pp. 247--267, 2001.

\bibitem{anderson2011reproducibility}
J.~S. Anderson, M.~A. Ferguson, M.~Lopez-Larson, and D.~Yurgelun-Todd,
  ``Reproducibility of single-subject functional connectivity measurements,''
  \emph{American journal of neuroradiology}, vol.~32, no.~3, pp. 548--555,
  2011.

\bibitem{tzourio2002automated}
N.~Tzourio-Mazoyer \emph{et~al.}, ``Automated anatomical labeling of
  activations in spm using a macroscopic anatomical parcellation of the mni mri
  single-subject brain,'' \emph{Neuroimage}, vol.~15, no.~1, pp. 273--289,
  2002.

\bibitem{vecchi2010nino}
G.~A. Vecchi and A.~T. Wittenberg, ``El ni{\~n}o and our future climate: where
  do we stand?'' \emph{Wiley Interdisciplinary Reviews: Climate Change},
  vol.~1, no.~2, pp. 260--270, 2010.

\bibitem{tachibana2011parietal}
A.~Tachibana, J.~A. Noah \emph{et~al.}, ``Parietal and temporal activity during
  a multimodal dance video game: an fnirs study,'' \emph{Neuroscience letters},
  vol. 503, no.~2, pp. 125--130, 2011.

\bibitem{saygin2008retinotopy}
A.~P. Saygin and M.~I. Sereno, ``Retinotopy and attention in human occipital,
  temporal, parietal, and frontal cortex,'' \emph{Cerebral Cortex}, vol.~18,
  no.~9, pp. 2158--2168, 2008.

\end{thebibliography}

\newpage
\section*{Supplemental Material}
\subsection{Lemmas and Proofs}
\begin{lemma}\label{lem:ld_range}
For any set $S$ of standardized variables, $\sigma_S \in [0,1]$.
\end{lemma}
\begin{proof}
Since we assumed all the variables to have unit variance, the sum of all the eigenvalues of the covariance matrix is exactly equal to $k$, the number of variables. Therefore, 
\begin{align}\label{Eq1}
     \lambda_{min}\leq 1.
 \end{align}
  Hence, 
  \begin{align}
    \sigma_S = 1 -\lambda_{min}\geq 0
  \end{align}
 $$\lambda_{min}\leq 1$$ and $$\sigma_S\geq 0.$$ 
 Further, since $\Sigma$ is positive semi-definite, we get
 \begin{align}\label{Eq0}
   \lambda_{min}\geq 0.
 \end{align}
 Also, by definition, the linear depedence $\sigma_S$ is given by 
 \begin{align}\label{Eqeig}
\sigma_S = 1 - \lambda_{min}
\end{align}
 Therefore, combining Eqs~(\ref{Eqeig}), (\ref{Eq1}) and (\ref{Eq0}), we get $$\sigma_S \in [0,1].$$. 

\end{proof}

\begin{lemma}\label{lem:ld}
The linear dependence of a set $S$ is always less than or equal to that of its supersets.
\end{lemma}
 \begin{proof}
Consider a superset $S' = S \cup X_{k+1}$ of S and let the corresponding data matrix be $X' = [X_1,X_2,X_3,...,X_{k+1}]$. Let $l^{*} = [l^{*}_1,l^{*}_2,...l^{*}_k]^T$, where $||l^{*}||_2 = 1$, denote the linear coefficients corresponding to the LVNLC $Z^{*}_S$. Consider $Z^{'}_{S'} = \mathbf{X'} l^{'}$, where $l^{'} = [l^{*}_1,l^{*}_2,...l^{*}_k,0]^T$. Then clearly,
 \begin{align*}
 var(Z^{'}_{S'}) = var(Z^{*}_S)  
 \end{align*}
 Let  $Z^{*}_{S'}$ be the LVNLC for set $S'$. Then, by definition of LVNLC and the above equality, we get 
 \begin{align*}
 var(Z^{*}_{S'}) \leq var(Z^{'}_{S'}) = var(Z^{*}_S)
\end{align*}
 Therefore, 
 \begin{align*}
     \sigma_{S'} = 1 - var(Z^{*}_{S'}) \geq 1 - var(Z^{*}_{S}) =  \sigma_S
 \end{align*}
\end{proof}

\begin{lemma}\label{lem:pseudo}
A set $S$ is a \textbf{negative-equivalent clique} \textit{iff} it can be partitioned into two negative cliques $S_1$ and $S_2$ such that the all the cross correlations between members of $S_1$ and $S_2$ are non-negative.
\end{lemma}
\begin{proof}

Consider a negative clique $S = \{X_1,X_2,...,X_k\}$, where all the pairwise correlations are negative. Let $S$ be partitioned arbitrarily into non-empty subsets $S_1$ and $S_2$. Hence, $S_1$ and $S_2$ will also be negative cliques. Without loss of generality, let us flip the signs of all variables in $S_1$ to obtain $S_1' = \{-X_1,-X_2,...,-X_j\}$. Due to this transformation, all the cross-correlations between $S_1'$ and $S_2$ will flip the signs and hence will become non-negative. Furthermore, every pairwise correlation within $S_1'$ remain negative, since $corr(X,Y) = corr(-X,-Y)$. Also correlations within $S_2$ remain negative. As a result, the negative equivalent set $S'$ can be partitioned into two negative cliques $S_1'$ and $S_2$ with all non-negative cross-correlations between them. Since $S'$ was generated using an arbitrary partitioning of an arbitrary negative clique $S$, this holds true for all negative equivalent cliques. 

To prove converse, consider a set $S$ that can be partitioned into two negative cliques $S_1$ and $S_2$ such that all pairwise cross-correlations between $S_1$ and $S_2$ are non-negative. Now, let us flip the signs of all variables in any one of the two partitions say $S_1$ (w.l.o.g) to form $S_1'$. Then by similar arguments as above, all pairwise correlations within $S_1'$ and $S_2$ will remain negative, while the cross correlations between $S_1$ and $S_2$ flip their signs from positive to negative. Hence, all the pairwise correlations in the transformed set $S_1'\cup S_2$ are negative, which implies that it is a negative clique, and hence, $S$ is a negative equivalent clique. 
\end{proof}

\begin{lemma}\label{lem:pseudo2}
A set $S$ is a \textbf{pseudo negative-equivalent clique} \textit{iff} it can be partitioned into two pseudo negative cliques $S_1$ and $S_2$ such that the all the cross correlations between members of $S_1$ and $S_2$ are pseudo non-negative.
\end{lemma}
\begin{proof}
Consider a pseudo negative clique $S = \{X_1,X_2,...,X_k\}$, where all the pairwise correlations are $\leq\rho$. Let $S$ be partitioned arbitrarily into non-empty subsets $S_1$ and $S_2$. Hence, $S_1$ and $S_2$ will also be pseudo negative cliques. Without loss of generality, let us flip the signs of all variables in $S_1$ to obtain $S_1' = \{-X_1,-X_2,...,-X_j\}$. Due to this transformation, all the cross-correlations between $S_1'$ and $S_2$ will flip the signs and hence will become $\geq-\rho$.  Furthermore, every pairwise correlation within $S_1'$ remain $\leq \rho$, since $corr(X,Y) = corr(-X,-Y)$. Similarly, correlations within $S_2$ remain $\leq \rho$. As a result, the pseudo negative equivalent set $S'$ can be partitioned into two pseudo negative cliques $S_1'$ and $S_2$ with all  cross-correlations between them $\geq -\rho$. Since $S'$ was generated using an arbitrary partitioning of an arbitrary pseudo negative clique $S$, this holds true for all pseudo negative-equivalent cliques. 

To prove converse, consider a set $S$ that can be partitioned into two pseudo negative cliques $S_1$ and $S_2$ such that all pairwise cross-correlations between $S_1$ and $S_2$ are $\geq -\rho$. Now, let us flip the signs of all variables in any one of the two partitions say $S_1$ (w.l.o.g) to form $S_1'$. Then by similar arguments as above, all pairwise correlations within $S_1'$ and $S_2$ will remain $\leq \rho$, while the cross correlations between $S_1$ and $S_2$ flip their signs and will now $\leq \rho$. Hence, all the pairwise correlations in the transformed set $S_1'\cup S_2$ are  $\leq \rho$, which implies that it is a pseudo negative clique, and hence, $S$ is a pseudo negative-equivalent clique. 
\end{proof}
 
\begin{algorithm}[!h]
\footnotesize
  \caption{CoMEt (\textbf{C}lique Based \textbf{M}ultipole S\textbf{e}arch)}\label{Algo:Comet}
  \begin{algorithmic}[1]
  \Statex{\textbf{Input} Dataset:$\mathcal{D}$, Parameters: $\delta$}
  \Statex{\textbf{Output} set $U$ of maximal multipoles with linear gain $\geq \delta$.}
  \State{$C \gets $ FIND PROMISING CANDIDATES($D$)}
  \For{each clique $S$ in $C$}
    \State{$U \gets$ GET MULTIPOLES FROM CANDIDATE($S$)}
  \EndFor{}
  
  \State $U \gets $ REMOVE DUPLICATES \& NON-MAXIMALS($U$)
  \State \textbf{return} $U$
  \end{algorithmic}
\end{algorithm}

\subsection{Location-based to Region-based SLP Time Series}\label{LocToReg}
Relationships in climate datasets are preferably studied between regions (sets of spatially contiguous locations) as opposed to individual locations because of i) spatial autocorrelation, due to which locations in a spatial neighborhood have highly similar time series that will lead to discovery of redundant relationships, and ii) the fact that the relationships studied across regions are likely to be more reliable and stable over time. Therefore, in this work, we converted the location-based Sea Level Pressure (SLP) time series dataset into a region-based time series dataset using the procedure described in Algorithm~\ref{Algo:GetReg}. Specifically, treating each of 10512 locations $l_i$ as a centre, we grew a spatially contiguous region $R_i$ around it by including up to the top 50 locations that were most strongly correlated to $l_i$ and spatially contiguous. For each region, a representative time series is obtained as the normalized area-weighted average time series of all the included locations. Note that there would be a high degree of redundancy among the  regions due to spatial autocorrelation and mutual spatial-overlaps among them. Therefore, we followed steps elaborated in lines 7-16 of Algorithm~\ref{Algo:GetReg}, to obtain a subset $D$ of representative time series of 171 non-redundant regions that are non-redundant and cover entire globe.

\begin{algorithm}[!h]
\footnotesize
  \caption{GET NON-REDUNDANT REGION TIME SERIES}\label{Algo:GetReg}
  \begin{algorithmic}[1]
  \Statex{\textbf{Input:} A set $L$ of 10512 locations}
  \Statex{\textbf{Output} set $D$ of representative time series of non-redundant regions}
  \State{$U_R,U_T = \phi$}
  \For{ each location $l_i \in L$}
    \State{$R_i \gets $ a set of top 50 locations most strongly correlated with $l_i$ and are spatially contiguous}
    \State{$T_i \gets$ Area-weighted mean of time series of all locations included in $R_i$}
    \State{Insert $R_i$ into $U_R$ and $T_i$ into $U_T$ }
  \EndFor
  
  \Statex{\textbf{Count number of redundant regions for every region}}
  \For{each region $R_i \in U_R$}
    \State{$A \gets $ Set of all regions $R_j$ s.t. $corr(T_i,T_j) \geq 0.8$ }
    \State{$RedunCount(R_i) \gets$ $|A|$}
  \EndFor
  \Statex{\textbf{Remove redundant regions}}
  \State{$D = \phi$}
  \While{$U_R \neq \phi$}
    \State{$R_k \gets \argmax\limits{R_i \in U_R}(RedunCount(R_i))$}
    \State{Insert $T_k$ into $D$}
    \State{Remove all regions from $U_R$ that are redundant to $R_k$}
  \EndWhile
  \end{algorithmic}
\end{algorithm}

\subsection{Time Complexity Analysis}
The following discusses the time complexity analysis of the the three parts of Algorithm~\ref{Algo:Comet}. As is common with many pattern finding techniques, such as frequent pattern mining, the algorithm is inherently exponential. However, as with association analysis, adjusting the parameter settings, $\sigma_S$ and $\Delta\sigma_S$, can make the pattern search quite tractable. 

\textbf{Part 1 (line 1 in Algorithm~\ref{Algo:Comet})}: This part performs two tasks: i) constructing a graph, and ii) enumerating cliques from the graph. Graph construction requires computation of correlation matrix, which $O(N^2 T)$ in time, where N is the number of time series and $T$ is the length of the time series. Clique enumeration is known to be an NP-complete problem. The worst case time complexity of enumerating cliques in an m-vertex graph is given by $O(3^{m/3})$. Hence, the total time complexity of this submodule will be $ O(N^2T) + O(3^{m/3})$. 

\textbf{Part 2 (lines 2-4 in Algorithm~\ref{Algo:Comet})}: This part uses finds multipoles from all the promising candidates. For a given promising candidate $S$ containing $k$ variables, computing linear dependence and linear gain requires one to do eigenvalue decomposition on a $k \times k$ matrix as well as $k$ submatrices of size $(k-1) \times (k-1)$.  Time complexity of standard algorithms for eigenvalue decomposition is $O(k^3)$. Thus, total time complexity of  evaluating linear dependence and linear gain for  a set $S$ of  $k$ variables is $O(k^4)$. In the worst case, for each candidate of size $k$, one would have to check all possible subsets of sizes $[3,k]$ to check if they form a multipole. Total such subsets would be less than $2^k$. Hence, total cost for a candidate of size $k$ would be upper bounded by $O(2^{k}k^{4})$. In the worst case, the total number of promising candidates outputted will be $O(3^{m/3})$, where $m$ is the vertex-connectivity of correlation graph in dataset. Thus, worst case complexity of this part will be $O(3^{m/3}2^{l}l^{4})$.  Here $l$ is the size of the largest candidate.

\textbf{Part-3 (line 5 of Algorithm~\ref{Algo:Comet}):} For a given collection of $p$ multipoles, total time complexity of removing non-maximals could be performed in $O(p2^{l})$ in time using hash data structure, where $l$ is the largest size of the multipole. In the worst case, the number of multipoles would be proportional to the number of promising candidates obtained. Hence, the worst case time complexity of this submodule will be $O(3^{m/3}2^{l})$.

Thus, the overall time complexity of Algorithm~\ref{Algo:Comet} would be the sum of time complexity of all the three parts: $O(N^2T) + O(3^{m/3}) + O(3^{m/3}2^{l}l^{4})$. Here $N$ is the number of number time series, $T$ is the length of each time series, $m$ is the vertex-connectivity of the graph in which maximal cliques are enumerated, $l$ is the size of the largest clique obtained. 

\subsection{CoMetExtended is complete under certain scenarios} 
Under the assumption of correctness of our empirical observations described in Section 4.1 of the manuscript, we claim the following: \\
    
        \textbf{Claim:} \textit{For any given user-specified threshold on linear gain, there exists a range of values of parameter $\rho$ for which CoMEtExtended is guaranteed to perform a complete search of multipoles.} \\

\begin{figure*}[t]
    \centering
    \begin{subfigure}[t]{0.30\textwidth}
    \includegraphics[width=4.5cm,height=4.5cm]{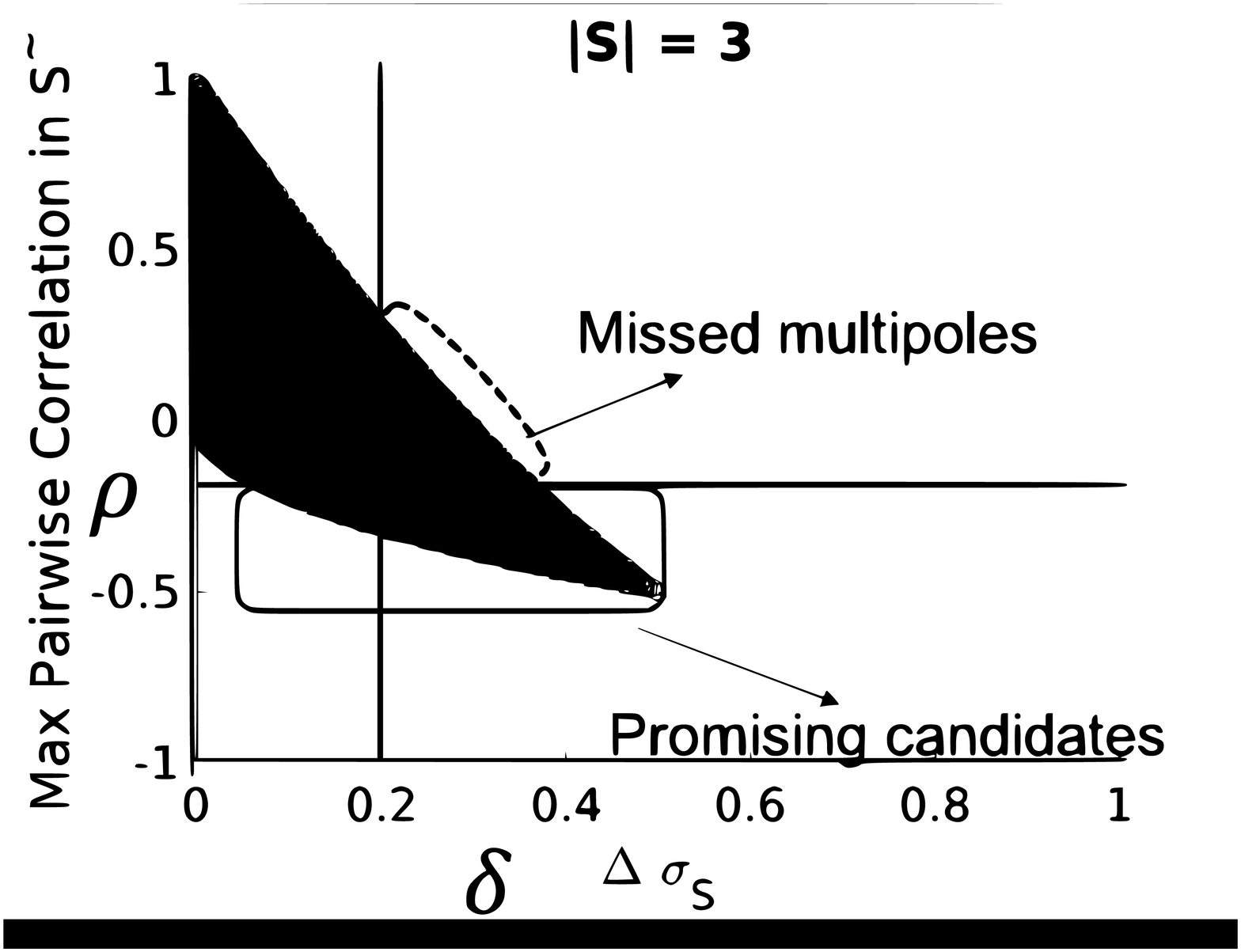}
    \centering
    \caption{$|S|=3$}
    \label{fig:rhodelta3}
    \end{subfigure}
    \centering
    \begin{subfigure}[t]{0.30\textwidth}
    \includegraphics[width=4.5cm,height=4.5cm]{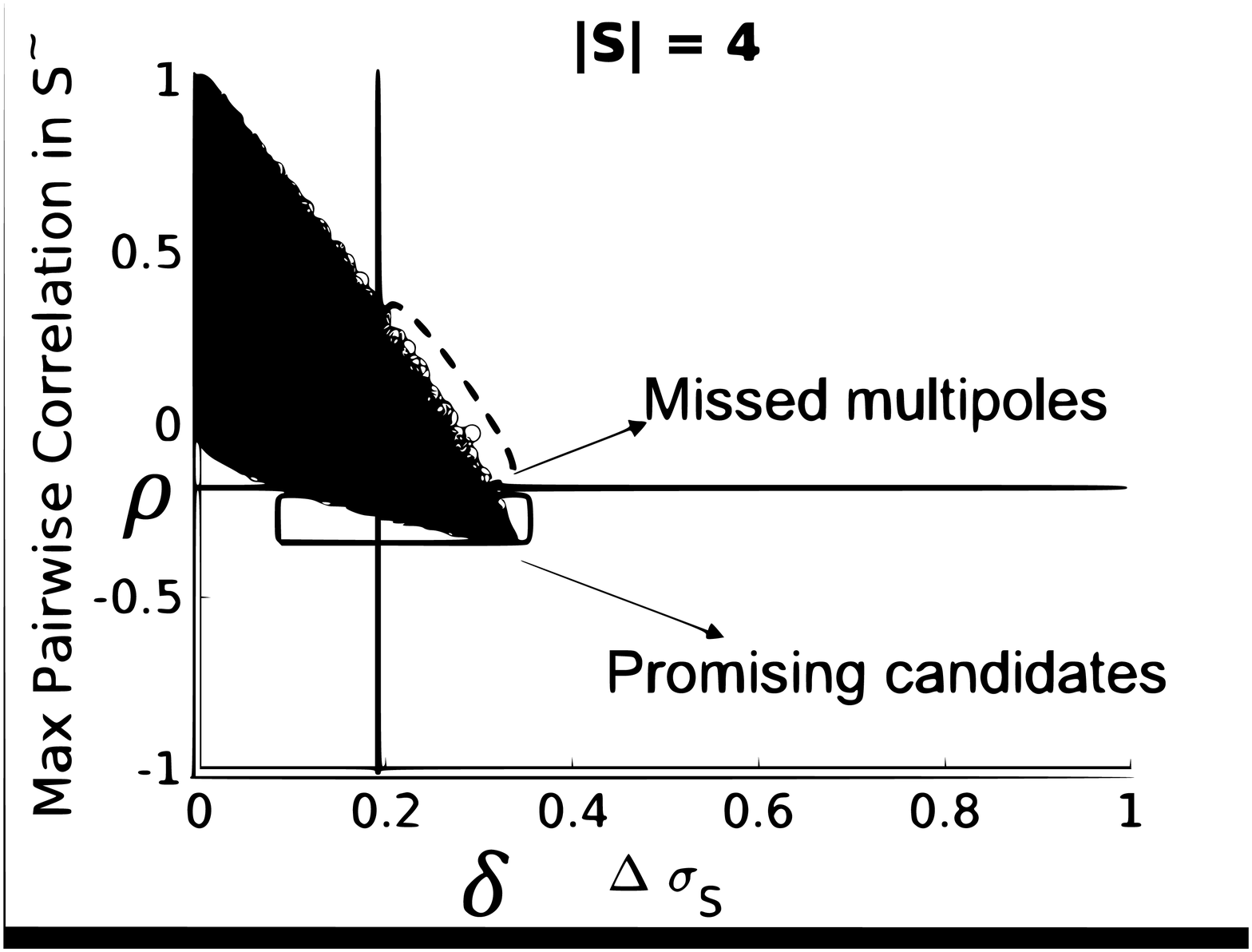}
    \centering
    \caption{$|S|=4$}
    \label{fig:rhodelta4}
    \end{subfigure}
    \centering
    \begin{subfigure}[t]{0.30\textwidth}
    \includegraphics[width=4.5cm,height=4.5cm]{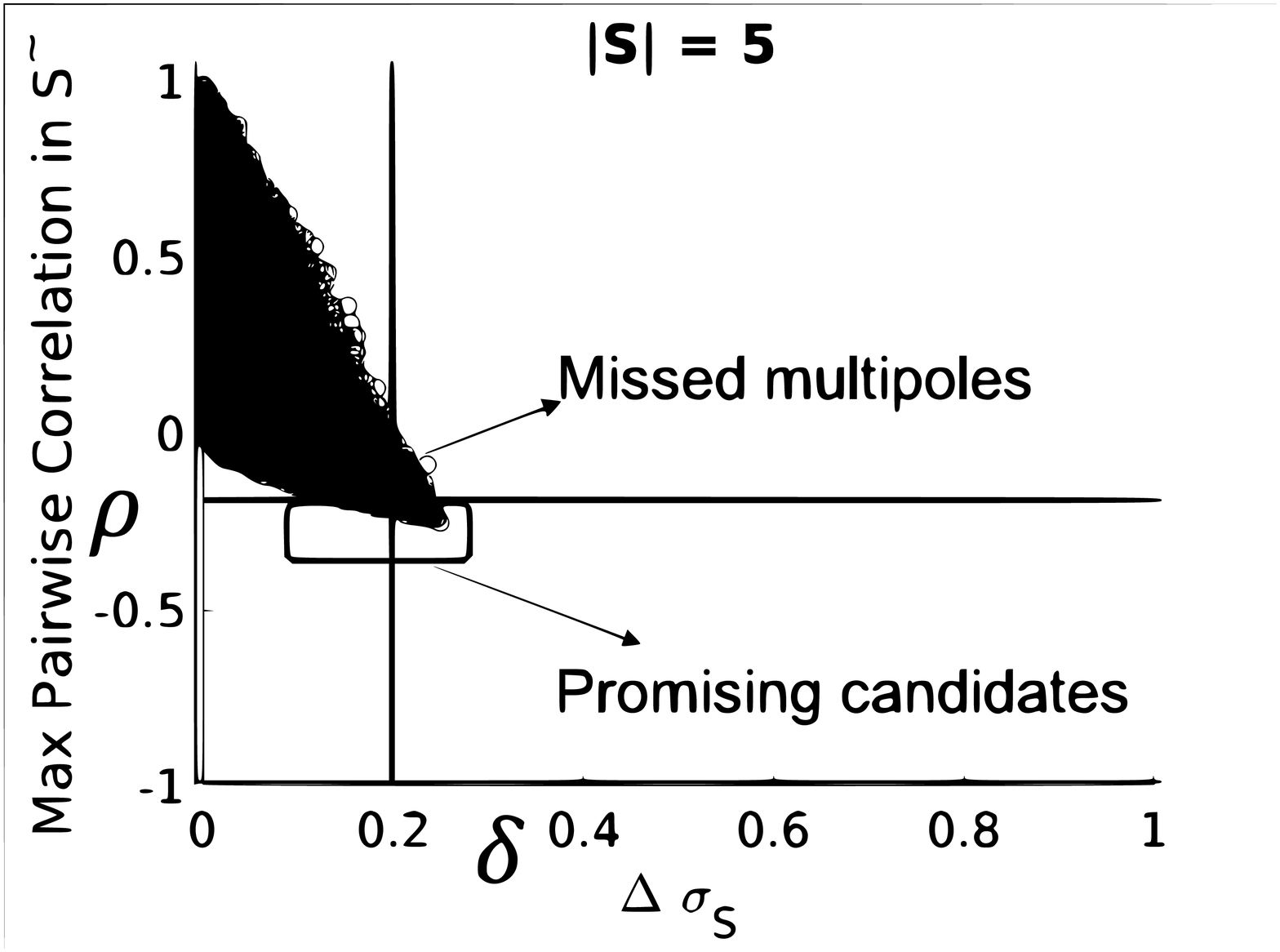}
    \centering
    \caption{$|S|=5$}
    \label{fig:rhodelta5}
    \end{subfigure}
    \caption{\textbf{} Empirical observations for correlation matrices of sizes $3\times3$, $4\times4$, and  $5\times5$ in (a), (b), and (c) respectively.}
    \label{fig:RhoDelta}
    \end{figure*}

    Such a range of values of $\rho$ could be determined by a simple procedure described as follows: 
    Figure~\ref{fig:RhoDelta} revisits the scatter plots of Figure 3 in the manuscript, where the X-axis indicates the linear gain of a set and the Y-axis indicates the largest pairwise correlation in one of its canonical versions called the `self-canceling' version, which has exactly the same linear gain and linear dependence as of the original set. Each scatter corresponds to a correlation matrix of size $3\times3$, $4\times4$, and $5\times5$ in Figures~\ref{fig:rhodelta3},\ref{fig:rhodelta4}, and \ref{fig:rhodelta5} respectively. 
    
    Note that for a given threshold $\delta$ on linear gain,  the multipoles of interest are located towards the right of the vertical red line as shown in all three plots of Figure~\ref{fig:RhoDelta}. In contrast, as described in Section~4.4 of the manuscript,  the search space of CoMEtExtended is determined by the value of $\rho$, and is restricted to sets that are located below the corresponding horizontal green line shown in plots of Figure~\ref{fig:RhoDelta}. Therefore, for a given combination of $\delta$ and $\rho$, all the multipoles lying in the missing zone (region to the right of red line and the top of green line) will be missed. Thus to guarantee completeness, we can choose any value of $\rho$ for which the missing zone is empty. Since higher values of $\rho$ lead to higher computational time, an ideal value of $\rho$ will be $\rho'$, the smallest one that will guarantee completeness. 
    
\begin{figure*}[h]
    \centering
    \begin{subfigure}[t]{0.3\textwidth}
    \includegraphics[width=4.5cm,height=4.5cm]{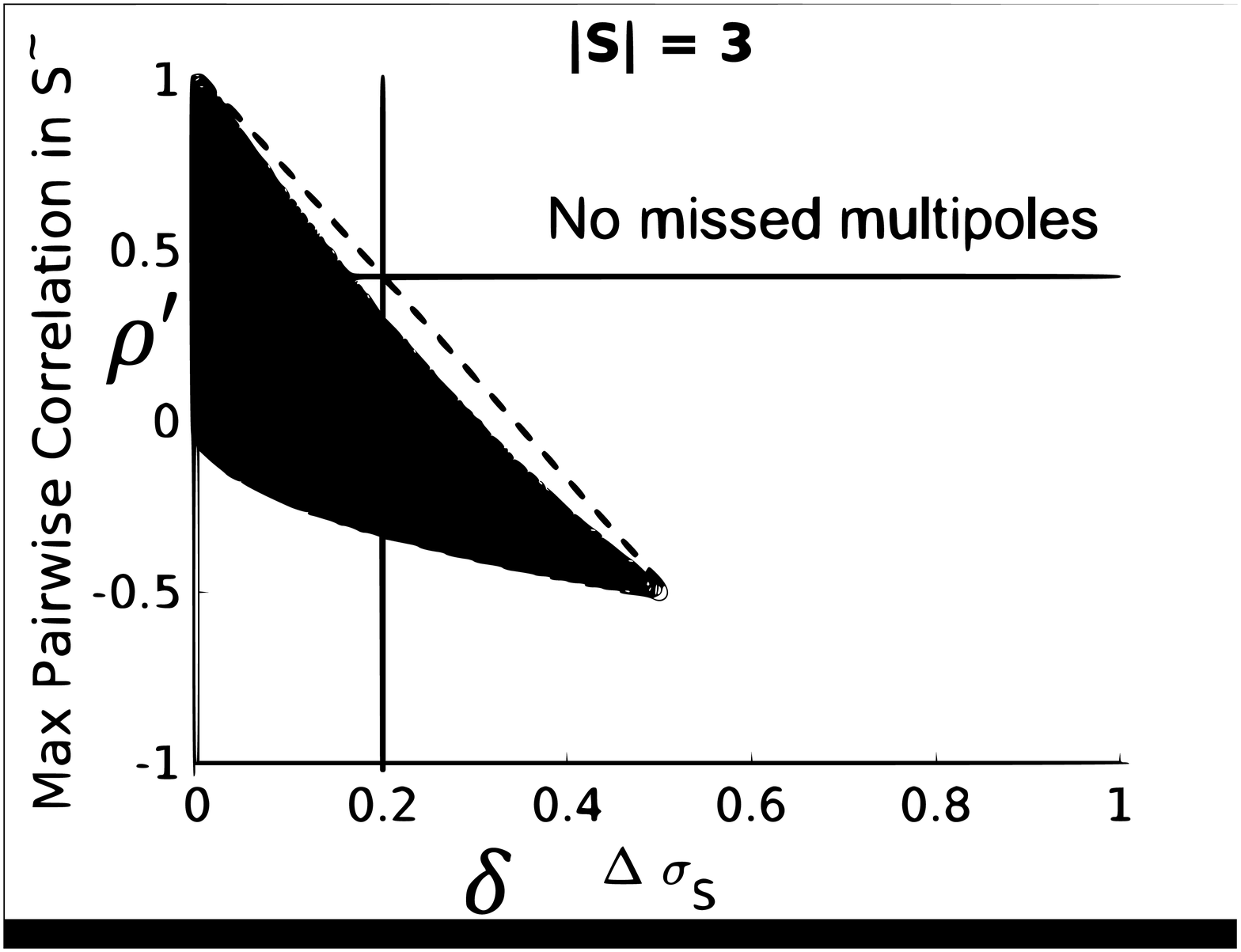}
    \centering
    \caption{$|S|=3$}
    \label{fig:blackline3}
    \end{subfigure}
    \centering
    \begin{subfigure}[t]{0.3\textwidth}
    \includegraphics[width=4.5cm,height=4.5cm]{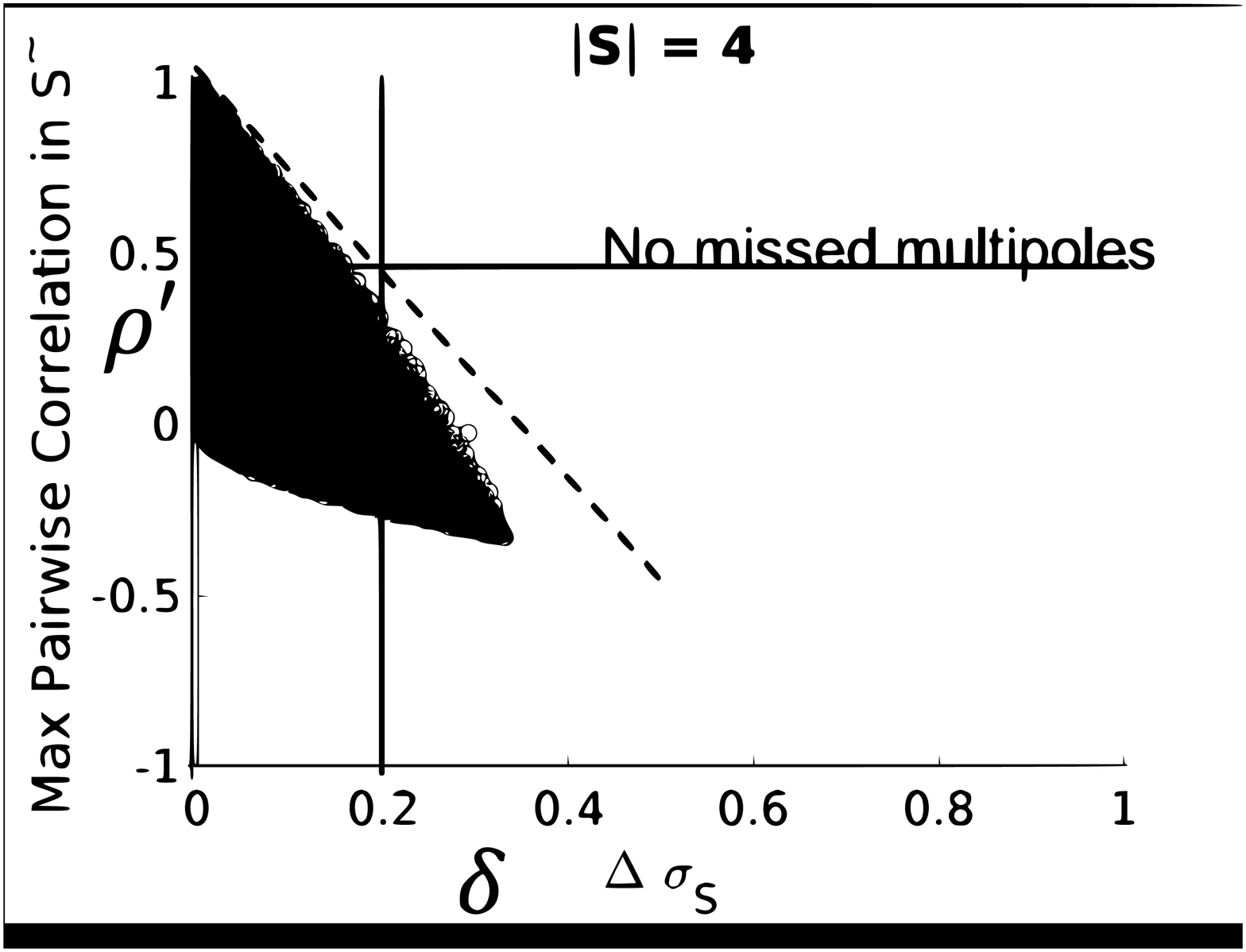}
    \centering
    \caption{$|S|=4$}
    \label{fig:blackline4}
    \end{subfigure}
    \centering
    \begin{subfigure}[t]{0.3\textwidth}
    \includegraphics[width=4.5cm,height=4.5cm]{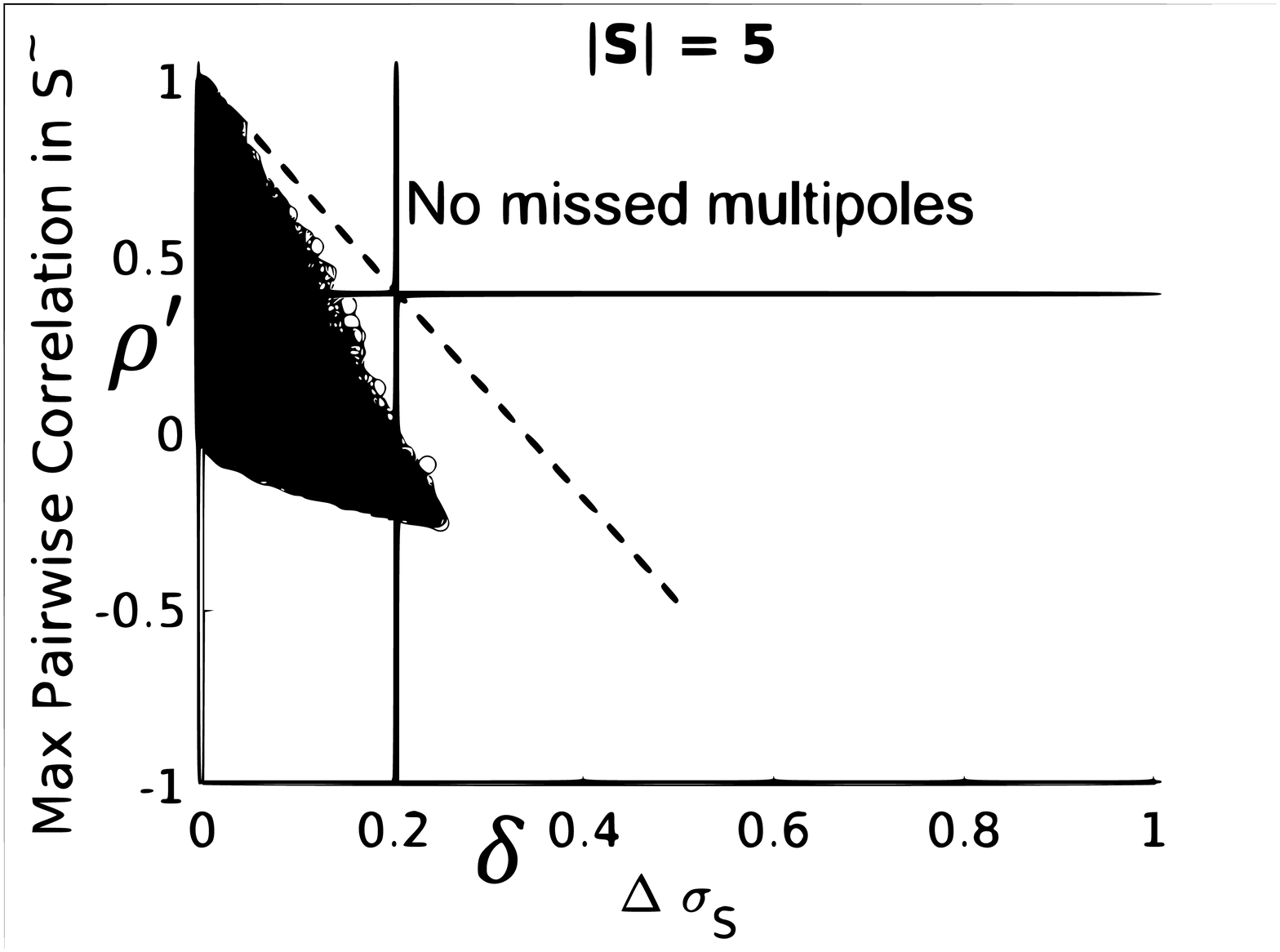}
    \centering
    \caption{$|S|=5$}
    \label{fig:blackline5}
    \end{subfigure}
    \caption{\textbf{} Estimating empirical upper bound on the lowest value of $\rho$ that guarantees completeness. The three plots in (a), (b), and (c) indicate empirical observations for correlation matrices of size $3\times3$, $4\times4$, and  $5\times5$ respectively.}
    \label{fig:BlackLine}
    \end{figure*}

    An empirical upper bound on $\rho'$ for a given threshold $\delta$ could be obtained as the ordinate of the intersection point of the vertical red line (that indicates threshold on linear gain) and a curve that marks an upper boundary of the distributions of correlation matrices of all sizes. An example of such a curve is a straight black-dash line as shown in plots of Figure~\ref{fig:BlackLine}, given by the equation $y+3x = 1$, and thus an upper bound on $\rho'$ can be given by $-3\delta + 1$.   For $\rho < \rho'$, CoMEtExtended will be approximately complete and the degree of completeness will vary across datasets.

\subsection{Evaluation on Large Synthetic Datasets}
 To further demonstrate the utility of our approach on larger datasets, we performed additional experiments on 4 synthetic datasets of sizes 10k,20k. 40k, and 100k time series of length 1000 timestamps. In each synthetic dataset of size $N$ time series, there are 200 time series corresponding to  66 multipoles of sizes 3 to 5, each forming a multipole with linear gain at least 0.1 and linear dependence at least 0.7, while the remaining $N-200$ time series are independently generated using a white noise Gaussian process. 
 
 The procedure to generate a synthetic multipole is based on cholesky decomposition of a correlation matrix that satisfies the given thresholds on linear dependence and linear gain. In detail, to generate a synthetic multipoles of size $k$, we randomly generate a correlation matrix $\Sigma$ of size $k\times k$ that shows linear dependence and gain to be at least 0.7 and 0.1 respectively. Next we perform a cholesky decomposition of the correlation matrix $\Sigma = LL^T$ to obtain the matrix $L$. Let $X$ denote a matrix of size $k\times N$, where each row is a time series of length $T$ generated using a white noise Gaussian processes. Then, a desired multipole of size $k$ could be obtained as $XL$ that will have $\Sigma$ as their correlation matrix and thus would form a desired multipole.

Table~\ref{tab:syneval} shows the run-time (in seconds) and the fraction of inserted multipoles recovered (in \%) by CoMEtExtended  on all the four datasets for different values of parameter $\rho$ in range $[-0.15,-0.1]$. For all the datasets, our approach was able to recover all the 66 multipoles in less than 2 hours. Note that for all the datasets, both the run-time and completeness increases with increase in the value of $\rho$. This is expected because the parameter $\rho$ determines the trade-off between computational efficiency and completeness of the search.  At high negative values of $\rho$, the graph generated is extremely sparse and therefore, the clique enumeration problem is solved much faster. However, the completeness of the approach is also relatively low, especially if the threshold $\delta$ on linear gain is very high. On the other hand, as $\rho$ is set to higher values, fewer edges are  pruned which generates a denser graph for which the clique enumeration problem takes more time. Also notice that in this experiment, the degree of completeness of the search remains the same for all datasets for a given value of $\rho$. This is expected since all the datasets are built to have identical set of multipoles. However, this need not be true in general and the degree of completeness can vary across datasets for a given value of $\rho$.  


\begin{table}[h]
\footnotesize
\centering
\centering
\begin{tabular}{|c|c|c|c|c|}
\hline
 &\multicolumn{4}{|c|}{Run-time in seconds} \\
 &\multicolumn{4}{|c|}{ (fraction of inserted multipoles)} \\
 \cline{2-5}
\textbf{$\rho$} & \multicolumn{4}{|c|}{\textbf{\# of time series in dataset (N)}} \\\cline{2-5}
&\textbf{N=10000} &\textbf{N=20000} &\textbf{N=50000} &\textbf{N=100000} \\\hline
-0.15 &15(1.5\%) &64(1.5\%) &371(1.5\%) &2889(1.5\%) \\ \hline
-0.14 &15(4.5\%) &65(4.5\%) &397(4.5\%) &2944(4.5\%) \\ \hline
-0.13 &16(7.6\%) &66(7.6\%) &415(7.6\%) &3050(7.6\%) \\ \hline
-0.12 &17(24\%) &74(24\%) &420(24\%) &3112(24\%) \\ \hline
-0.11 &19(55\%) &76(55\%) &429(55\%) &5042(55\%) \\ \hline
-0.1 &32(100\%) &155(100\%) &855(100\%) &5227(100\%) \\ \hline

\end{tabular}
\caption{\textbf{Evaluation on Synthetic Data:} Computational time (in seconds) and fraction of multipoles recovered by CoMEtExtended at different values of parameter $\rho$ in four synthetic datasets of different sizes. Threshold parameters $\sigma$ and $\delta$ were set to 0.7 an 0.1 respectively.}
\label{tab:syneval}
\end{table}
\subsection{Evaluation Against Structure Learning Methods}\label{sec:GL}
Another class of methods that we discussed in related work include structure learning methods. These methods are designed to learn the structure of stochastic dependencies among variables in a dataset in the form of a graphical model called \emph{Markov network}. Markov network is a graph where each node represents a variable and follows the pairwise Markovian property, according to which it is independent of any non-neighboring node in the network conditioned on Markov blanket’, i.e. the set of all of its immediate neighboring nodes. Therefore, Markov networks are typically designed to infer the conditional independence between different subsets of variables using various statistical inference techniques. In contrast, the multipole patterns are defined to capture direct or indirect dependencies between different subsets of variables. Hence, there doesn't seem to be a principled approach to infer all multipoles from a Markov network. In the following, we describe an experiment to compare the performance of our proposed approach against a structure-learning method based- baseline in finding multipoles. 
        
The Markovian property of the Markov network implies that there exists direct dependencies between a node and its neighboring nodes. Thus, a potential approach to generate multipole candidates from a Markov network is to group every node with all of its neighboring nodes. Based on this idea, we designed a structure learning method-based baseline for finding multipoles in the data. In this baseline, we used \textit{Graphical LASSO}, a popular structure learning method to learn a sparse Markov network $G$ from the data.  For each node $v_i$ and its Markov Blanket $MB_i$ in $G$, we first ranked all the nodes in $MB_i$ based on decreasing order of their edge weights (edge with $v_i$), and then generated multiple multipole candidates by grouping $v_i$ with its top $k$ neighboring nodes for $k\in [2,|MB_i|]$, thereby obtaining $|MB_i|-1$ number of multipole candidates for node $v_i$. To further improve the results, we ran Graphical LASSO at multiple values of regularization parameter to obtain multiple Markov networks at different levels of sparsity, and then combined the multipole candidates obtained from all the networks into the single output. 

\begin{table}[h]
\footnotesize
\centering
\centering
\begin{tabular}{|c|c|c|c|}
\hline 
&Total Multipoles in &\multicolumn{2}{|c|}{Completeness} \\
 \cline{3-4}
\textbf{($\sigma$,$\delta$)} & Pseudo-complete set &SLB &CoMEtExtended \\
\hline
\textbf{(0.4,0.1)} &70150 &0.14\% &81\%, $\rho$=$0.01$\\ \hline 
\textbf{(0.4,0.15)} &6255 &0.5\% &96\%, $\rho$=$0.01$\\ \hline 
\textbf{(0.4,0.2)} &1264 &0.63 \% &99\%, $\rho$=$0.01$\\ \hline 
\textbf{(0.5,0.1)} &41126 &0.24\% &76\%, $\rho$=$0.01$\\ \hline 
\textbf{(0.5,0.15)} &3348 &0.93\% &92\%, $\rho$=$0.01$\\ \hline 
\textbf{(0.5,0.2)} &930 &0.86\% &99\%, $\rho$=$0.01$\\ \hline 
\textbf{(0.6,0.1)} &13743 &0.7\% &75.5\%, $\rho$=$0.03$\\ \hline 
\textbf{(0.6,0.15)} &1525 &2.03\% &85\%, $\rho$=$0.01$\\ \hline 
\textbf{(0.6,0.2)} &488 &1.64\% &98\%, $\rho$=$0.01$\\ \hline 
\end{tabular}
\caption{Completeness evaluation of CoMEtExtended against Structure Learning baseline (SLB) at different combinations of $\sigma$ and $\delta$ in the SLP dataset. The parameter $\rho$ in CoMEtExtended was set so as to keep the computational time under 90 minutes.}
\label{tab:compevalSLP}
\end{table}

\begin{table}[h]
\footnotesize
\centering
\centering
\begin{tabular}{|c|c|c|c|}
\hline 
&{\footnotesize Total Multipoles in} &\multicolumn{2}{|c|}{Completeness} \\
 \cline{3-4}
\textbf{($\sigma$,$\delta$)} & {\footnotesize Pseudo-complete set} &SLB &CoMEtExtended \\
\hline
\textbf{(0.4,0.1)} &15855 &0.04\% &71\%, $\rho$=$0.2$\\ \hline 
\textbf{(0.4,0.15)} &3019 &0.03\% &98\%, $\rho$=$0.2$\\ \hline 
\textbf{(0.4,0.2)} &716 &0\% &100\%, $\rho$=$0.2$ \\ \hline 
\textbf{(0.5,0.1)} &15258 &0.04\% &70\%, $\rho$=$0.2$\\ \hline 
\textbf{(0.5,0.15)} &2805 &0.04\% &98\%, $\rho$=$0.2$\\ \hline 
\textbf{(0.5,0.2)} &697 &0\% &100\%, $\rho$=$0.2$\\ \hline 
\textbf{(0.6,0.1)} &13721 &0.04\% &82\%, $\rho$=$0.25$\\ \hline 
\textbf{(0.6,0.15)} &2172 &0.05\% &97\%, $\rho$=$0.2$\\ \hline 
\textbf{(0.6,0.2)} &547 &0\% &100\%, $\rho$=$0.2$ \\ \hline 
\end{tabular}
\caption{Same as Table~\ref{tab:compevalSLP}, but in the fMRI dataset.}
\label{tab:compevalfMRI}
\end{table}
    
Tables~\ref{tab:compevalSLP} and \ref{tab:compevalfMRI} show the fraction of total multipoles (pseudo-complete set) detected by this approach for different choices of thresholds on linear dependence and linear gain in SLP and fMRI dataset respectively. The regularization parameter $\lambda$ in the objective function of Graphical LASSO was varied from 0.1 to 1 in steps of 0.001 for SLP dataset, and 0.01 to 0.4 in steps of 0.001 for fMRI dataset. As can be seen, the fraction of multipoles discovered by SLB is minimal and less than about 2 \% for all the choices of thresholds. The poor performance of SLB baseline highlights the challenges in inferring multipoles from a Markov network, which renders structure learning based approaches unsuitable for finding multipoles in the data.
\end{document}